\documentclass{bmvc2k}

\title{Learning visual representations for transfer learning by suppressing texture}

\addauthor{Shlok Mishra}{shlokm@umd.edu}{1}
\addauthor{Anshul Shah}{ashah95@jhu.edu}{2}
\addauthor{Ankan Bansal}{ankan@umd.edu}{1}
\addauthor{Janit Anjaria}{janit@cs.umd.edu}{1}
\addauthor{Jonghyun Choi}{jc@yonsei.ac.kr}{3}
\addauthor{Abhinav Shrivastava}{abhinav@cs.umd.edu}{1}
\addauthor{Abhishek Sharma}{abhisharayiya@gmail.com}{4}
\addauthor{David Jacobs}{djacobs@umiacs.umd.edu}{1}

\addinstitution{
 University of Maryland, College Park
}
\addinstitution{
Johns Hopkins University
}
\addinstitution{
Yonsei University
}
\addinstitution{
Axogyan AI
}
\usepackage{times}
\usepackage{epsfig}
\usepackage{graphicx}
\usepackage{amsmath}
\usepackage{amssymb}

\usepackage{rotating}
\usepackage{colortbl}
\usepackage{arydshln}
\usepackage{booktabs}
\usepackage{enumitem}
\usepackage{balance}
\usepackage{combelow}
\usepackage{tabularx}
\definecolor{Gray}{gray}{0.85}
\definecolor{darkgreen}{rgb}{0.0, 0.8, 0.0}
\definecolor{Lightgray}{gray}{0.90} 
\newcolumntype{a}{>{\columncolor{Lightgray}}c}
\usepackage{graphicx}
\usepackage{wrapfig}
\usepackage[stable]{footmisc}

\usepackage{multirow}   
\usepackage{makecell}
\usepackage{algorithm}
\usepackage[noend]{algpseudocode}
\usepackage{algorithmicx,eqparbox,array}

\usepackage{xcolor,colortbl}

\usepackage{url}

\runninghead{Mishra et.al }{Improving transfer learning by suppressing texture}

\begin{document}

\maketitle
\begin{abstract}
Recent literature has shown that features obtained from supervised training of CNNs may over-emphasize texture rather than encoding high-level information.  In self-supervised learning, in particular, texture as a low-level cue may provide shortcuts that prevent the network from learning higher-level representations.  We hypothesize that retaining more edge information and suppressing texture can help in alleviating these problems. To this end, we propose to use a simple classical idea based on anisotropic diffusion to augment training using images with suppressed texture. 
We empirically show that our method achieves improved results on image classification with five diverse datasets in both supervised or self-supervised learning tasks such as MoCoV2 and Dense-CL. 
Our method is particularly effective for transfer learning tasks, and we observed improved performance on twelve transfer learning datasets. 
The large improvements (up to 11.49\%) on the Sketch-ImageNet and Synthetic-DTD datasets, and
additional visual analyses of saliency maps suggest that our approach helps in learning better representations that transfer well to downstream tasks. We show that our method is simple to implement and can be integrated into various computer vision tasks easily.
\end{abstract}

\section{Introduction}
Deep convolutional neural networks (CNNs) learn powerful visual features that have resulted in significant improvements on many computer vision tasks such as semantic segmentation \cite{Shelhamer_2017}, object recognition \cite{NIPS2012_4824}, and object detection \cite{Ren2015FasterRT}. 
However, CNNs often fail to generalize well across datasets under domain-shift due to varied lighting, sensor resolution, spectral-response etc. One of the reasons for this poor generalization is CNNs' over-reliance on low-level cues like texture \cite{Geirhos2018ImageNettrainedCA}.

These low-level cues and texture biases have been identified as grave challenges to various learning paradigms ranging from supervised learning \cite{wiel2019approximating, Geirhos2018ImageNettrainedCA,ringer2019texture} to self-supervised learning (SSL) \cite{Noroozi2016UnsupervisedLO, Noroozi_2018_CVPR,Doersch2015UnsupervisedVR, Caron2018DeepCF,Devlin2019BERTPO}. We propose to use classical tools to suppress texture in images as a form of data augmentation to encourage deep neural networks to focus more on learning representations that are less dependent on textural cues.  We use the Perona-Malik non-linear diffusion method \cite{Perona1990ScaleSpaceAE}, robust Anisotropic diffusion \cite{Black1998RobustAD}, and Bilateral filtering \cite{tomasi1998bilateral} to augment our training data.  
These methods suppress texture while retaining structure by preserving boundaries.

\begin{figure*}[t]
    \centering
    \includegraphics[width=0.7\linewidth]{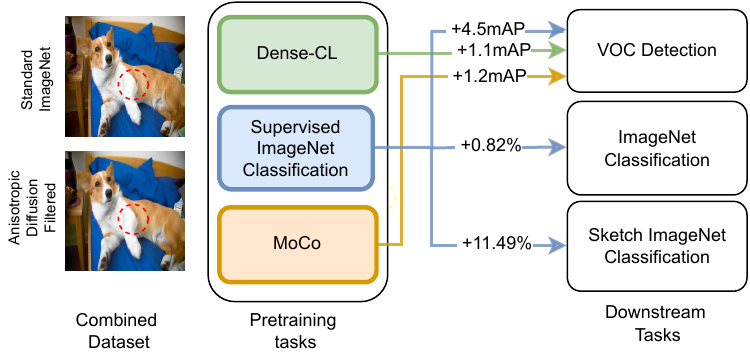}
    \caption{An overview of our approach. We propose to augment the ImageNet dataset by adding with Anisotropic diffused images. The use of this augmentation helps the network rely less on texture information and increases performance in several settings.}
    \label{fig:teaser}
\end{figure*}

Our work is inspired by the observation that  ImageNet pre-trained models fail to generalize well across datasets \cite{Geirhos2018ImageNettrainedCA,recht2019imagenet}, due to over-reliance on texture and low-level features. Stylized-ImageNet \cite{Geirhos2018ImageNettrainedCA}  attempted to modify the texture of images using style-transfer to render images in the style of randomly selected paintings from the Kaggle paintings dataset.
However, this approach offers little control over exactly which cues are removed from the image. The resulting images sometimes retain texture and distort the original shape. Stylized-ImageNet especially doesn't work well in the case of SSL since the network learns the texture and uses those textures to solve the SSL tasks (see Tab \ref{tab:MocoV2 Experiments} Row1.  
In our approach (Fig. \ref{fig:teaser}), we suppress the texture instead of modifying it. We empirically show that this helps in learning better higher-level representations and works better than CNN-based stylized augmentation. We compare our approach with Gaussian blur augmentation, recently used in \cite{chen2020simple, chen2020improved}, and show that Anisotropic-filtering for texture suppression is better because isotropic Gaussian blur can potentially suppress edges and other higher-level semantic information as well. Our proposed method works well in self-supervised and supervised learning tasks, and we outperform both Gaussian blurring and Stylized-ImageNet in both settings. 

Anisotropic-filtering is simple to implement and can be integrated easily in various computer vision-based methods. In the case of supervised learning, we pre-train on ImageNet, and test on twelve different datasets including ImageNet, Pascal VOC \cite{Everingham2009ThePV}, Synthetic-DTD \cite{Newell_2020}, CIFAR 100  \cite{hendrycks2019selfsupervised}, Sketch ImageNet \cite{Wang2019LearningRG}, etc. For self-supervised setting, we use two learning frameworks: Dense-CL \cite{wang2020DenseCL}, and MoCoV2 \cite{chen2020improved} and pre-train on ImageNet and COCO. Our texture-suppressing augmentation consistently outperforms MoCoV2 and Dense-CL, which uses Gaussian blurring, on transfer learning experiments on VOC classification, detection, segmentation benchmarks, and also on classification tasks for other transfer learning datasets, including  DTD \cite{cimpoi14describing}, Cars \cite{KrauseStarkDengFei-Fei_3DRR2013}, Aircraft \cite{maji13fine-grained}, etc. With the help of qualitative and quantitative analysis, we show that our model is less reliant on high-frequency information in images and is more robust to common corruptions on datasets like ImageNet-C \cite{Hendrycks2019BenchmarkingNN}, and CIFAR-100 \cite{hendrycks2019selfsupervised}. Our model also learns better shape bias than a Standard-ImageNet pretrained model and is also more confident in making correct predictions.  Overall, we achieve significant improvements on several benchmarks: 
 \begin{itemize}
\setlength\itemsep{1pt}
     \item In a set of \textbf{twelve} diverse datasets, our method exhibits substantial improvements (as high as $+11.49\%$ on Sketch ImageNet and $+10.41\%$ on the DTD dataset) in learning visual representations across domains.
     \item We also get improvements in the same domain visual recognition tasks on ImageNet validation (+0.82\%), and on label corruption task \cite{hendrycks2019selfsupervised}.
    \item We achieve improved results in self-supervised learning on image classification transfer learning tasks and  on VOC detection.
\end{itemize}

\begin{figure*}
\centering
\includegraphics[ width=\linewidth]{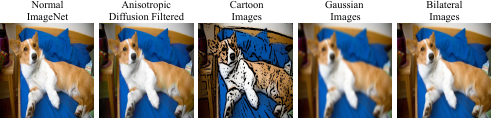}
\caption{Four different methods for reducing texture in images.  
}
\label{fig:anisotropic}
\end{figure*}
\section{Related Work}
In this section, we review relevant methods that aim to remove texture cues from images to reduce the dependency of CNNs on low-level cues. Since we also experiment with the application of our method in self-supervised learning, we review recent work in this area as well.

\textbf{Reliance on Low-Level Texture Cues.} 
Recent studies have highlighted that deep CNNs can leverage low-level texture information for classification on the ImageNet dataset. Contrary to popular belief that CNNs capture shape information of objects using hierarchical representations \cite{lecun2015deep}, the work in \cite{Geirhos2018ImageNettrainedCA} revealed that CNNs trained on ImageNet are more biased towards texture than shape information. This dependency on texture not only affects generalization, but it can also limit the performance of CNNs on emerging real-world use-cases like few-shot image classification \cite{ringer2019texture}.  
\cite{wiel2019approximating} showed that a bag of CNNs with limited receptive field in the original image can \emph{still} lead to excellent image classification performance. Intuitively, a small receptive field forces the CNNs to heavily rely on local cues vs learning hierarchical shape representations. This evidence strongly suggests that texture alone can yield competitive performance on ImageNet, and the fact that it's relatively easier to learn vs hierarchical features may explain deep CNNs' bias towards texture. \\ 
To reduce reliance on texture, Stylized-ImageNet  \cite{Geirhos2018ImageNettrainedCA} modified the ImageNet images into different styles taken from the Kaggle Painter by Numbers dataset. While trying to remove texture, this approach could also significantly affect the shape. Also, there isn't an explicit control over the amount of removed texture. Moreover, this method may not be directly applicable to self-supervised learning because the fixed number of possible texture patterns result in images with strong low-level visual cues resulting in shortcuts. We show that the accuracy on downstream tasks, when MoCoV2 and Jigsaw are trained with Stylized-ImageNet, decreases dramatically (Table 1 Supplementary). 
Some of the recent methods inspired by Stylized-ImageNet \cite{Geirhos2018ImageNettrainedCA} that try to learn better shape representation \cite{mummadi2021does,islam2021shape,li2021shapetexture} face similar issues. Info-Dropout\cite{shi2020informative} tries to learn shape information by using dropout based methods which zeros out neurons by a high probability if the input patch contains less self-information.
\\
On the other hand, we use Perona-Malik's anisotropic diffusion \cite{Perona1990ScaleSpaceAE} and bilateral filtering \cite{tomasi1998bilateral} as ways of suppressing texture in images. These methods remove texture without degrading the edge information. Consequently, the shape information of the objects is better preserved. Also, these methods provide finer control over the level of texture suppression. Suppressing the texture in training images forces CNN to build representations that put less emphasis on texture. We show that such data augmentation can lead to performance improvements in both supervised and self-supervised settings. 
We also distinguish our work from other data augmentation strategies like Auto-Augment \cite{Cubuk2018AutoAugmentLA} which uses Reinforcement Learning to automatically search for improved data augmentation policies and introduces Patch Gaussian Augmentation, which allows the network to interpolate between robustness and accuracy \cite{Lopes2019ImprovingRW}. 


\textbf{Self-Supervised Learning.}
To demonstrate the importance of removing texture in the self-supervised setting, we consider two pretext tasks. The first pretext task is Jigsaw \cite{Noroozi2016UnsupervisedLO}, a patch-based self-supervised learning method that falls under the umbrella of visual permutation learning \cite{Cruz2017DeepPermNetVP,Cruz2018VisualPL} . 
Some of the most recent self-supervised methods are contrastive learning based methods \cite{he2019momentum,Caron2018DeepCF,hnaff2019dataefficient,Hjelm2018LearningDR,misra2019selfsupervised,ge2021robust,chen2020simple,chen2020improved,Caron2019UnsupervisedPO,NEURIPS2020_4c2e5eaa,NEURIPS2020_f7cade80,mishra2021objectaware,ge2021robust,Shah_2022_WACV, lee2021imix,wang2020DenseCL, caron2021unsupervised}. In
\cite{Caron2018DeepCF}, the authors have proposed using contrastive losses on patches, where they learn representations by predicting representations of one patch from another.
In MoCo \cite{he2019momentum}, a dynamic dictionary is built as a queue along with a moving average encoder. Every image will be used as a positive sample for a query based on a jittered version of the image. The queue will contain a batch of negative samples for the contrastive losses.  
MoCo has two encoder networks. The momentum encoder has weights updated through backpropagation on the contrastive loss and a momentum update. In MoCoV2, Gaussian blur and linear projection layers were added that further improve the representations.    
MoCo and MoCoV2 have shown competitive results on ImageNet classification and have outperformed supervised pre-trained counterparts on seven detection/segmentation tasks, including PASCAL VOC \cite{Everingham2009ThePV}, and COCO \cite{Lin2014MicrosoftCC}.

\textbf{Transfer Learning.}
Transfer learning is one of the most important problems in computer vision due to difficulty in collecting large datasets across all domains. In this work, we discuss transfer learning in the context of ImageNet. A lot of early datasets were shown to be too small to generalize well to other datasets \cite{Torralba2011UnbiasedLA}. Following this, many new large-scale datasets were released \cite{imagenet_cvpr09,Lin2014MicrosoftCC}, which are believed to transfer better. However, recent results show that these datasets do not generalize well in all cases \cite{recht2019imagenet,Kornblith2019DoBI}. \cite{Kornblith2019DoBI} showed that ImageNet features generally transfer well, but not to fine-grained tasks. We show results of transfer learning on some of the datasets that were used by \cite{Kornblith2019DoBI}. 
\section{Methods}

CNN-based classifiers have been shown to exploit textures rather than shapes for classification \cite{Geirhos2018ImageNettrainedCA,wiel2019approximating}. 
We aim to reduce the prominence of texture in images and thus encourage networks trained to learn representations that capture better higher-level representations.

\textbf{Gaussian Blur.} Gaussian blurring is one of the most popular smoothing methods in computer vision, and it has been recently proposed as data augmentation for SSL \cite{chen2020simple,chen2020improved}. However, along with low-level texture, Gaussian filtering also blurs across boundaries, diminishing edges and structural information.  
 




\subsection{Anisotropic diffusion}
\label{subsec:anis} We propose to use  Anisotropic Diffusion Filters (ADF) ~\cite{Perona1990ScaleSpaceAE}, which keep the shape information coherent and only alter low-level texture.  Specifically, we use Perona-Malik diffusion \cite{Perona1990ScaleSpaceAE}. 
These filters smooth the texture without degrading the edges and boundaries. Intuitively, this will encourage the network to extract high-level semantic features from the input patches.  




Perona-Malik diffusion smooths the image using the differential diffusion equation:
\begin{align}
\frac{\partial I}{\partial t} &=c(x, y, t) \Delta I+\nabla c \cdot \nabla I\\    
&c(x, y, t)  =e^{-(\|\nabla I(x, y, t)\| / K)^{2}}
\end{align}
where $I$ is the image, $t$ is the time of evolution, $\Delta$ is the Laplacian operator, $K$ controls sensitivity to edges, $\nabla$ is gradient and $(x,y)$ is a location in the image. 
%
%
The amount of smoothing is modulated by the magnitude of the gradient in the image through $c$ the diffusion coefficient. The larger the gradient, the smaller the smoothing at that location. Therefore, after applying Anisotropic diffusion, we obtain images with blurred regions, but edges are still prominent. Fig.~\ref{fig:anisotropic} shows some examples of the application of the filter.  
Note that ADF reduces the texture in the image without replacing it, the domain gap between images is not large, while in the case of Stylized ImageNet, the domain shift will be large.
Recently, there has been some work on removing textures using deep learning as well \cite{Xu2014ScaleInvariantCN,Liu2016LearningRF,lu2018deep}. We find, though, that fast and simple classical methods work well on most tasks.
For all our experiments, we create a  dataset `Anisotropic ImageNet' by adding ADF filtered ImageNet images to the standard ImageNet dataset. We also experiment with training an Image-to-Image translation model Pix2Pix \cite{isola2018imagetoimage} to suppress texture. We train the model to produce images that are similar in style to anisotropic diffusion. Pix2pix model also helps in capturing different variations of texture, e.g., the amount of smoothing,  according to the target task. Details of the Pix2Pix based approach are mentioned in the supplementary. 

\subsection{Other texture suppressing methods}
We also experiment with a few other texture removing methods like robust Anisotropic diffusion \cite{Black1998RobustAD}, Bilateral filtering \cite{tomasi1998bilateral}, and Cartoonization. However, empirically we find that the most simple Anisotropic diffusion method has the best results as discussed in Section \ref{Superisved_learning}. We will discuss these other texture removing methods briefly next.

\textbf{Bilateral Filtering:} \cite{Tomasi1998BilateralFF} is an efficient method of anisotropic diffusion.  
In Gaussian filtering, each pixel is replaced by an average of neighbouring pixels, weighted by their spatial distance. 
Bilateral Filtering is its extension in which weights also depend on photometric distance.  
This also limits smoothing across edges, in which nearby pixels have quite different intensities.

\textbf{Cartoonization:} A more extreme method of limiting texture is to create cartoon images.  To convert an image into a cartoonish image, we first apply bilateral filtering to reduce the image's colour palette. In the second step, we convert the actual image to grayscale and apply a median filter to reduce noise in the grayscale image. 
After this, we create an edge mask from the greyscale image using adaptive thresholding. 
Finally, we combine these two images to produce cartoonish looking images (see Fig. \ref{fig:anisotropic}).

\section{Experiments}

We start by briefly describing the datasets used in our experiments. We then show the effectiveness of ADF for supervised and self-supervised learning. We find that ADF is particularly effective when there is a domain shift, supporting our hypothesis that variation in texture is a significant effect of domain shift.  
The effect is larger when we transfer from ImageNet to datasets such as Sketch Imagenet \cite{Wang2019LearningRG}, and Synthetic-DTD \cite{Newell_2020}, where the domain shift is larger. Our method is also able to outperform Stylized-ImageNet \cite{Geirhos2018ImageNettrainedCA} and Gaussian Blur \cite{chen2020simple}.

\textbf{Datasets.}
In all experiments, we use the ImageNet training set as the source of our training data.
For object detection and semantic segmentation, we evaluate on Pascal VOC 2007 and VOC 2012. For label corruption, we evaluate on CIFAR100.
When the downstream task is classification we evaluate on Synthetic-DTD \cite{Newell_2020}, Sketch-ImageNet \cite{Wang2019LearningRG}, Birds \cite{WahCUB_200_2011},
Aircraft \cite{maji13fine-grained}, Stanford Dogs\cite{Khosla2012NovelDF}, Stanford Cars  \cite{KrauseStarkDengFei-Fei_3DRR2013},DTD \cite{cimpoi14describing}, ImageNet-C \cite{Hendrycks2019BenchmarkingNN},  and the  ImageNet validation dataset.

\textbf{Experimental Details.} 
For SSL we build on MoCoV2 \cite{chen2020improved} and Dense-CL \cite{wang2020DenseCL}.
For supervised learning, we use the ResNet50 \cite{He2015} model, closely following \cite{Geirhos2018ImageNettrainedCA}.
After training on Anisotropic ImageNet, we fine-tune our model on the standard ImageNet training set following the procedure of \cite{Geirhos2018ImageNettrainedCA}. 
We set the conduction coefficient ($K$) of Anisotropic Diffusion to 20, and a total of 20 iterations are used. We use MedPy implementation. 
All other hyper-parameters are described in the supplementary material.

\subsection{Self-Supervision for Transfer Learning}
\label{sec:mocov2}
We first experiment with Anisotropic ImageNet on Self-Supverised methods.
We have doubled the number of images (Anisotropic images + normal images) as compared to normal ImageNet. So for a fair comparison, we only train our methods for half the number of epochs compared to training with just ImageNet.  
We then fine-tune the network pre-trained on the Anisotropic ImageNet  for the downstream tasks, including image classification, object detection, and semantic segmentation on PASCAL VOC, and other transfer learning datasets. Since, we are removing low-level cues from the images, we expect to see better results when transferring to different datasets. 

\textbf{MoCo V2.}
We evaluate our method with MoCo V2 \cite{chen2020improved} and Dense-CL \cite{wang2020DenseCL}, which are the state-of-the-art methods in SSL.  MoCoV2 and Dense-CL \cite{wang2020DenseCL}  used Gaussian blurring with 0.5 probability as data augmentation. In our case, we add Anisotropic diffusion on the images with 0.5 probability, and for the remaining 50\% of the images, we apply Gaussian blurring with 0.5 probability. So, in our setup every image has 0.5 probability of coming from Anisotropic ImageNet, 0.25 of Gaussian blurring, and 0.25 of being normal ImageNet. Also, the number of iterations on Anisotropic filtering is chosen randomly between 10 to 20. For object detection  starting from a MoCoV2 initialization, we train a Faster R-CNN \cite{Ren2015FasterRT} with C4-backbone, which is fine-tuned end-to-end.

We show improvements over MoCoV2 and Dense-CL for object detection on the VOC Dataset.
In the first setup, we show improvements on COCO-based evaluation metrics (i.e., $\text{AP}_{50}$,
$\text{AP}_{0.05:0.05:0.95}$,  $\text{AP}_{75}$) as shown in the first three columns of Table \ref{tab:MocoV2Experiments}, achieve competitive results. 
We also observe an improvement of $1.3$ mean IoU on semantic segmentation \cite{long2015fully} over MoCo V2 baseline and $1.1$ over Dense-CL baseline. We also see improved performances on Dense-CL when we pre-train on COCO dataset as well.
These results show that in the case of transfer learning, we improve across different datasets. More details can be found in the supplementary material. Our method is not bound to a particular pretext task and can be potentially added to any state-of-art method to achieve even further improvements. In the supplementary material, we show that our method leads to improvements with the Jigsaw \cite{Noroozi_2018_CVPR} task. 
\begin{table*}[t!]
\centering
\renewcommand{\tabcolsep}{6pt}
\renewcommand{\arraystretch}{1.1}
\caption{Comparison with MoCoV2 and Dense-CL in SSL. We note that using Anisotropic diffusion with improves performance on VOC detection and Semantic Segmentation (SS). We test on COCO-based metrics as used in \cite{chen2020improved}. We also improve performance over the baseline on the semantic segmentation (SS) task \cite{long2015fully}. Although we have only focussed on MoCoV2 and Dense-CL, our technique can potentially be extended to other state-of-art methods.}
\begin{tabular}{@{}lcccccc@{}}  
\toprule
Methods    & Dataset & $\text{AP}_{50}$ &
\text{AP}
& $\text{AP}_{75}$  & mIoU (SS) \\
\midrule
Stylized ImageNet  & & 43.5 & 28.80 & 33.7 & -\\
Supervised ImageNet     & & 81.6 & 54.2 & 59.8 &  59.8\\
\midrule
MoCo V2  \cite{chen2020improved}   & ImageNet & 82.4 & 57.0 & 63.6 & 67.5 \\
MoCo V2 Anistropic (Ours)    & ImageNet & \textbf{83.7} & \textbf{58.2} & \textbf{64.8}  & \textbf{67.8}\\
\midrule
Dense-CL \cite{wang2020DenseCL}   & ImageNet & 82.8 & 58.7 & 65.2 &  69.4 \\
Dense-CL \cite{wang2020DenseCL} Anistropic (Ours)   & ImageNet & \textbf{83.5} & \textbf{59.6} & \textbf{66.4} &  \textbf{70.5} \\
\midrule
Dense-CL CC \cite{wang2020DenseCL}   & COCO & 81.7 & 56.7 & 63.0 &  67.5 \\
Dense-CL CC Anistropic (Ours) \cite{wang2020DenseCL}   & COCO & \textbf{83.1} & \textbf{57.9} & \textbf{64.2} &  \textbf{68.6} \\
\bottomrule
\end{tabular}
\label{tab:MocoV2Experiments}
\end{table*}
\begin{table*}[t!]
\centering
\renewcommand{\tabcolsep}{6pt}
\renewcommand{\arraystretch}{1.1}
\caption{Transfer learning across different datasets. Note that our approach leads to improvements in both supervised and SSL set-up.}
\begin{tabular}{@{}lccccc@{}}  
\toprule
Dataset    & Aircraft \cite{maji13fine-grained} & Birds \cite{WahCUB_200_2011} & Dogs \cite{Khosla2012NovelDF} & Cars \cite{KrauseStarkDengFei-Fei_3DRR2013} & DTD \cite{cimpoi14describing} \\
\midrule
Supervised (Reproduced)    & 90.88 & 90.3 & 85.35 & 92.1 & 72.66 \\
\midrule
SimCLR \cite{chen2020simple} & 88.1 & \_ & \_ & 92.1 & 73.2  \\
BYOL \cite{grill2020bootstrap} & 88.1 & \_ & \_ & 91.7 & {76.2}  \\
MoCo V2  \cite{chen2020improved}   & 91.57 & 92.13 & 87.13 & 92.8 & 74.7  \\
MoCo V2 Anistropic (Ours)    & \textbf{92.71} & \textbf{93.29} & \textbf{88.81} & \textbf{94.3} & \textbf{76.3} \\
\bottomrule
\end{tabular}
\label{tab:MocoV2 Experiments}
\end{table*}

\begin{table}[t]
\centering
\renewcommand{\tabcolsep}{6pt}
\renewcommand{\arraystretch}{1.1}
\caption{Experiments with Sketch-ImageNet. Use of Anisotropic ImageNet shows that our method is better at capturing representation that are less dependent on texture. }
\begin{tabular}{@{}lcc@{}} 
\toprule
Method    & Top-1 Acc & Top-5 Acc \\
\midrule
ImageNet Baseline     & 13.00 & 26.24    \\
Stylized Baseline     & 16.36 & 31.56    \\
Pix2Pix-Anisotropic (Ours)  & \textbf{24.49} & \textbf{41.81}   \\
\bottomrule
\end{tabular}
\label{tab: Shape_Experiments}
\end{table}

These results suggest that training the network on the Anisotropic ImageNet dataset forces
it to learn better representations. This is consistent with our hypothesis that Anisotropic diffusion leads to smoothing of texture in images. This forces the network to be less reliant on lower-level information to solve the pretext task and, hence, learn representations that focus on higher-level concepts.  

\textbf{Experiments with Stylized ImageNet on MoCoV2 and Jigsaw.}
We now show experiments that indicate that, while effective in a supervised setting, Stylized ImageNet does not help with SSL. We train a model with MoCoV2 and Jigsaw as pretext tasks on the Stylized-ImageNet (SIN) dataset \cite{Geirhos2018ImageNettrainedCA} and fine-tune on the downstream tasks of object detection and image classification on PASCAL VOC. In Table \ref{tab:MocoV2 Experiments} (and Table 2 in supplementary), we show that there is a huge drop in performance.
One reason for this failure using the SIN dataset could be that the model is able to memorize the textures in the stylized images since it only has 79,434 styles. This is not a problem in the original fully-supervised setting where the authors used SIN for supervised image classification. In that case, the network can learn to ignore texture to discriminate between classes.

\subsection{Transfer Learning for Supervised Learning}
\label{Superisved_learning}
As shown in the last section, suppressing texture leads to performance improvements in the case of domain transfer with SSL. In this section, we also show improvements in supervised learning and domain transfer. In the case of supervised learning we also show results using Pix2Pix-Anistropic model.
\subsubsection{Across Domains}
We hypothesize that learning with texture bias is most harmful to domain transfer.
Thus, we first describe a challenging experimental setup for learning visual representation across domains.

\textbf{Sketch-ImageNet.}
\label{Sketch-ImageNet}
For a cross-domain supervised learning setup, we chose to use the Sketch-ImageNet dataset  \cite{Wang2019LearningRG}.
Sketch-ImageNet contains sketches collected by making Google image queries ``sketch of X'', where ``X'' is chosen from the standard class names of ImageNet.
The sketches have very little to no texture, so performance on Sketch-ImageNet is a strong indicator of how well the model can perform when much less texture is present. 
As shown in Table \ref{tab: Shape_Experiments}, the difference between the Pix2Pix-Anisotropic model and the baseline model is 11.49\% for Top-1 accuracy.  
This result implies that our model captures less dependent representations on texture than standard ImageNet and Stylized ImageNet.

\textbf{Other Datasets - Aircraft, Birds, Dogs, and Cars.} We further evaluate our method on image classification tasks using four different fine-grained classification datasets.
We also observe improvement on image classification across five datasets in Table \ref{tab:MocoV2 Experiments}.
This shows that in the case of domain shift, capturing higher level semantics helps in better transfer learning performance.

\textbf{Object Detection.} The biggest improvement we observe on transfer learning is on object detection on Faster-RCNN \cite{Ren2015FasterRT} as shown in Table \ref{tab: Stylized_Classification_experiments}. This improvement suggests that we are able to attend to more high-level semantics, which helps in transfer learning performance on object detection.
\begin{table*}
\centering
\caption{Comparison using different texture removing methods, with different hyper-parameters for Anisotropic diffusion methods. We observe that the most simple \cite{Perona1990ScaleSpaceAE} performs the best, and removing more texture from images does not improve performance.}
\label{tab: Stylized_Classification_experiments}
\begin{tabular}{@{}lcccc@{}}  
\toprule
Method    & \# Iterations & Top-1 Acc & Top-5 Acc & Object Detection \\
\midrule
Baseline Supervised & - & 76.13 & 92.98 & 70.7 \\
Stylized ImageNet \cite{Geirhos2018ImageNettrainedCA} &-& 76.72 & 93.27 & 75.1\\
\midrule
Perona Malik with Pix2Pix \cite{Perona1990ScaleSpaceAE}  & 20 & \textbf{76.95} & \textbf{93.36} & \textbf{75.21} \\
Perona Malik \cite{Perona1990ScaleSpaceAE}  & 20 & 76.71 & 93.26 & 74.37 \\
Perona Malik \cite{Perona1990ScaleSpaceAE}  & 50 & 76.32 & 92.96 & 73.80 \\
Robust AD \cite{Black1998RobustAD}              & 20 & 76.58 & 92.96 & 73.33 \\
Robust AD \cite{Black1998RobustAD}              & 50 & 76.64 & 93.09 & 73.57 \\
Gaussian Blur                               & - & 76.21 & 92.64 & 73.26 \\
Cartoon ImageNet                            & - & 76.22 & 93.12 & 72.31   \\
Bilateral ImageNet                        & - & 75.99 & 92.90 & 71.34 \\

\bottomrule
\end{tabular}
\end{table*}
\subsubsection{Same Domain}

\textbf{ImageNet:}
In Table \ref{tab: Stylized_Classification_experiments}, we show results using Anisotropic ImageNet for supervised classification.  
We observe that Anisotropic ImageNet improves performance in both ImageNet classification and object detection. We also use Pix2Pix model for learning to suppress texture. Pix2Pix model converts a normal image to texture suppressed image. For Gaussian blurring experiments, we closely follow \cite{chen2020improved} and add a Gaussian blur operator with variance from 10 to 20 and train in a similar manner to Stylized ImageNet \cite{Geirhos2018ImageNettrainedCA}.
We can see that our proposed Pix2Pix Anisotropic ImageNet performs better than both Stylized ImageNet and  Gaussian blurring.
Hence, blurring the image completely without respecting boundaries and edges, or distorting the shape information by style transfer does not improve performance.

\textbf{Different Texture Removing Methods.}
We also provide results using different texture removing methods and different hyper-parameters for Anisotropic diffusion in Table \ref{tab: Stylized_Classification_experiments}. 
We observe that as we increase the number of iterations and remove more and more texture from images, performance starts to degrade, possibly due to the difference that comes in the data distribution after removing texture information. The most simple texture removing method \cite{Perona1990ScaleSpaceAE} has the best results.

\section{Analysis}
We now show qualitatively and quantitatively how our model is less dependent on texture information.
\subsection{Synthetic-DTD Dataset:}
\label{DTD}
To better demonstrate the effectiveness of less texture dependent representations, we used the dataset introduced by \cite{Newell_2020}.  This dataset provides four variations in images:  texture, color, lighting, and viewpoint.  This dataset is created by taking 47 different textures from DTD dataset\cite{Newell_2020} and applying them to a 3D dataset of 10 classes, called ShapeNet\cite{chang2015shapenet} to yield the same object rendered with different view-points and multiple textures.  It contains 480,000 training images and 72,000 testing images.
In this dataset, we made sure that texture information during training and testing are completely different. So, the texture is not a cue when we use this dataset. 
We evaluate our  Pix2Pix-Anisotropic model on this dataset and compare against the baseline normal ImageNet model. 
The Pix2Pix-Anisotropic model achieves a performance boost of 10.41\% in classification which suggests that we are indeed able to learn texture agnostic feature representations.

\subsection{Experiments on testing shape bias:} 
 To show the low reliance of our model on texture and greater reliance on high-level features like shape, we use the Geirhos Style-Transfer (GST) dataset \cite{Geirhos2018ImageNettrainedCA}. It consists of 1,248 images of 16 classes from ImageNet with shape and texture coming from different classes. This dataset teases apart the importance of shape vs texture for CNNs. We observe an improvement of 1.02\% on identifying the class representing the shape over a ImageNet pretrained Resent-50. This shows that we can successfully reduce the reliance on texture.

\section{Conclusion}
\label{sec:discussion}
We propose to help a CNN focus on high-level cues instead of relying on texture by augmenting the ImageNet dataset with images filtered with Anisotropic diffusion, in which texture information is suppressed. 
Empirical results suggest that using the proposed data augmentation for pretraining self-supervised models and for training supervised models gives improvements across ten diverse datasets. Noticeably, the 11.4\% improvement while testing the supervised model on Sketch ImageNet suggests that the network is capturing more higher-level representations than the models trained on ImageNet alone.

{\footnotesize
\section*{Acknowledgement}
SM was supported in part by, by the US Defense Advanced Research Projects Agency (DARPA) Semantic Forensics (SemaFor) Program under grant HR001120C0124. Any opinions, findings, and conclusions or recommendations expressed in this material are those of the author and do not necessarily reflect the views of the DARPA.
AS was supported by an ONR MURI grant N00014-20-1-2787.
JC was partly supported by the NRF grant (No.2022R1A2C4002300), IITP grant (No.2020-0-01361-003, AI Graduate School Program (Yonsei University), 10\%, No.2021-0-02068, AI Innovation Hub, 10\%) funded by the Korea government (MSIT).
}

\bibliography{egbib}
\end{document}


\maketitle

\section{Pix2Pix for suppressing texture}
\begin{figure}[h]
\includegraphics[width=\linewidth]{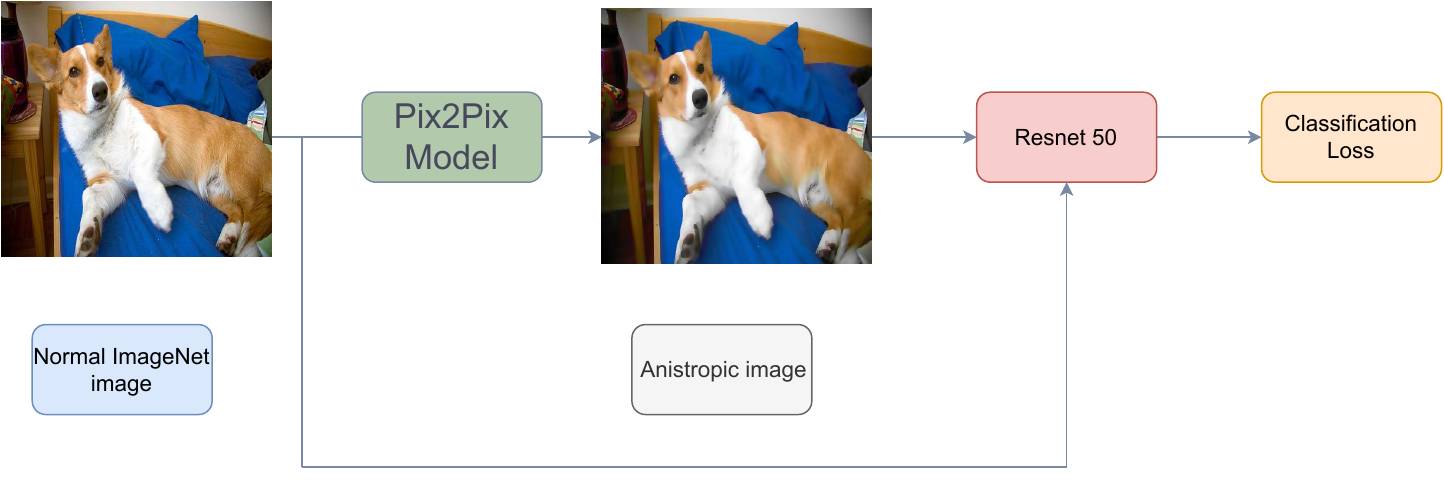}
\caption{The figure shows pipeline for training Pix2Pix model with the classification loss. The input to the Pix2Pix network is normal ImageNet image and output is Anisotropic image. Both of these images are used further for the classification task. }
\label{fig:pix2pix}
\end{figure}

In addition to using classical tools, we also apply recent Image-to-Image translation models like Pix2Pix \cite{isola2018imagetoimage} to suppress texture. We train pix2pix model to produce images that are similar in style to anisotropic diffusion. Pix2pix model also helps in capturing different variations of texture, e.g., the amount of smoothing,  according to the target task. The source domain for Pix2Pix model is normal ImageNet images, and the target domain is texture suppressed images. We only apply Pix2Pix model on Perona-Malik diffusion \cite{Perona1990ScaleSpaceAE}, which is our best performing anisotropic diffusion method. After adding Pix2Pix models and doing an end-to-end training for ImageNet classification, we are able to achieve better results and outperform Stylzied-ImageNet on \texttt{top-1} accuracy. We call this model  Pix2Pix-Anistropic model. 
For MoCoV2 we train using ten different anisotropic diffusion variations where the iterations are randomly picked between 10 to 20. Since we are already able to capture these variations and also due to the huge compute requirements and time to train SSL and Pix2Pix models, we only show these results in the supervised learning setting and not for MoCoV2. 
In Fig \ref{fig:pix2pix} we show our complete pipeline of training pix2pix model. We pretrain the pix2pix network on ImageNet dataset for 8 epochs and use this network for the end to end classification.

\section{Results on Medical imaging task:}
We also show  improved results on CheXpert and Chest14 datasets in Table \ref{2D} (last row). We see that our approach leads to an improvement on this challenging task. This shows the general nature of our approach and its applicability to medical imaging.

\begin{table*}[!b]
	\centering
	\caption{2D tasks: pretraining using Chest14 or CheXpert. We build on top of \cite{Zhou2021PreservationalLI} and add anistropic diffused images in the same setup as \cite{Zhou2021PreservationalLI}. We see improved performances for both CheXpert and Chest14 datasets.}
	{
    \resizebox{0.8\linewidth}{!}{ 
	\begin{tabular}{l|c|c|c|c|c|c|c|c|c|c|c|c}
	\toprule
	\multirow{}{Method} & \multicolumn{5}{c|}{Chest14$\rightarrow$Chest14} & \multicolumn{7}{c}{CheXpert$\rightarrow$Chest14} \\ \cline{2-13}
	& 9.5:0.5 & 9:1 & 8:2 & 7:3 & 6:4 & 10\% & 20\% & 30\% & 40\% & 50\% & 60\% & 100\% \\
	\hline
	TS & 61.8 & 68.1 & 71.5 & 73.4 & 75.4 & 68.1 & 71.5 & 73.4 & 75.4 & 77.5 & 79.1  & 80.9\\
	\hline
	IN & 70.5 & 73.6 & 75.3 & 76.9 & 78.0 & 73.5 & 76.3 & 78.4 & 79.0 & 79.5 & 79.7 & 81.0 \\
	\hline
	MG & 66.4 & 70.0 & 73.9 & 76.1 & 77.3 & 70.1 & 73.9 & 75.5 & 76.5 & 77.6 & 79.3 & 80.8 \\
	\hline
	SG & 66.5 & 70.2 & 74.3 & 76.7 & 77.6 & 69.7 & 73.8 & 75.6 & 77.3 & 77.3 & 79.6 & 81.3 \\
	\hline
	C2L & 71.7 & 74.1 & 76.4 & 77.5 & 79.0 & 73.1 & 77.0 & 78.5 & 79.1 & 79.8 & 80.2 & 81.5 \\
	\hline
	PCRL & {74.1} & {76.2} & {78.8} & {79.0} & {79.9} & {75.8} & {77.6} & {79.8} & {80.8} & {81.2} & {81.7} & {83.1} \\
	\hline
	PCRL + Anisotropic & \textbf{75.8} & \textbf{77.5} & \textbf{80.3} & \textbf{80.1} & \textbf{80.5} & \textbf{76.9} & \textbf{78.1} & \textbf{80.7} & \textbf{82.2} & \textbf{82.2} & \textbf{82.9} & \textbf{84.2} \\
	\bottomrule
	\end{tabular}}
	\label{2D}}	
\end{table*}

\section{Analysis Continued}
We now show few more experiments which show qualitatively and quantitatively how our model is less dependent on texture information.

\subsection{Visual Analysis:}
We now visually analyze the results by the saliency maps, which are produced by different networks. We use GradCam \cite{Selvaraju2016GradCAMVE} to calculate the saliency maps. In Fig \ref{fig:correct_incorrect}, we show the saliency maps produced by networks trained using the combined dataset and the original ImageNet dataset. 
We observe that Pix2Pix-Anisotropic model has saliency maps that spread out over a bigger area and that include the outlines of the objects. This suggests that it attends less to texture and more to overall holistic shape. In contrast, ImageNet trained models have narrower saliency maps that miss the overall shape and focus on localized regions.
In Fig.~\ref{fig:correct_incorrect}(a-e), we present the examples where the Pix2Pix-Anisotropic model gives the correct prediction, and the ImageNet model fails.
For example in Fig.~\ref{fig:correct_incorrect}(e), we observe that the network trained on ImageNet alone is not focusing on the whole bird and is only focusing on the body to make the decision; whereas the one trained with Pix2Pix-Anisotropic ImageNet is focusing on complete bird to make a decision. We include more saliency maps on Sketch-ImageNet, and cases where ImageNet trained models are correct and our model fails in the supplementary material.

\begin{figure*}[t!]
\centering
\resizebox{1\linewidth}{!}{
\begin{tabular}{cccccc}
\\
{\rotatebox{90}{\parbox{0.15\linewidth}{\centering Anisotropic\\Correct}}}&
\includegraphics[width=0.18\linewidth]{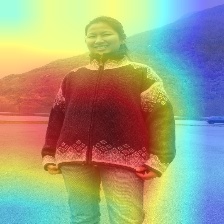}&
\includegraphics[width=0.18\linewidth]{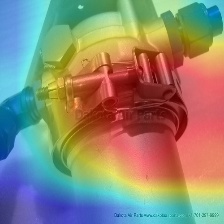}&
\includegraphics[width=0.18\linewidth]{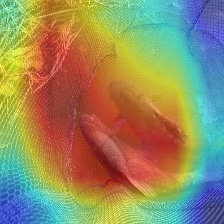}&
\includegraphics[width=0.18\linewidth]{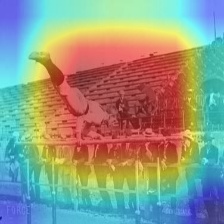}&
\includegraphics[width=0.18\linewidth]{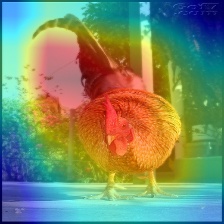}\\
{\rotatebox{90}{\parbox{0.15\linewidth}{\centering ImageNet\\Incorrect}}}&
\includegraphics[width=0.18\linewidth]{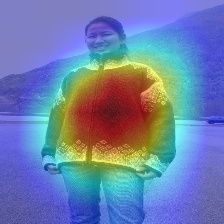}&
\includegraphics[width=0.18\linewidth]{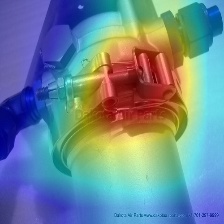}&
\includegraphics[width=0.18\linewidth]{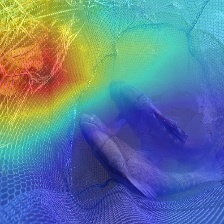}&
\includegraphics[width=0.18\linewidth]{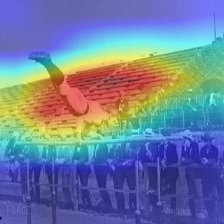}&
\includegraphics[width=0.18\linewidth]{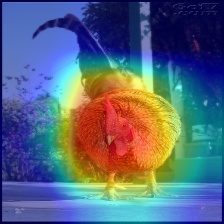}
\\
&(a)&(b)&(c)&(d)&(e)\\
\\

\end{tabular}
}
\caption{Saliency maps using GradCam. The text on the left of the row indicates whether the Anisotropic model or ImageNet model was used. The figure shows the saliency maps where the Anisotropic model gave correct predictions, and the ImageNet model gave wrong predictions. The failure of the ImageNet model might be due to it not attending to the whole object.}
  \label{fig:correct_incorrect}
\end{figure*}

\subsection{Results after removing high-frequency components:}
\begin{figure}[t]
    \centering
    \includegraphics[width=0.8\linewidth]{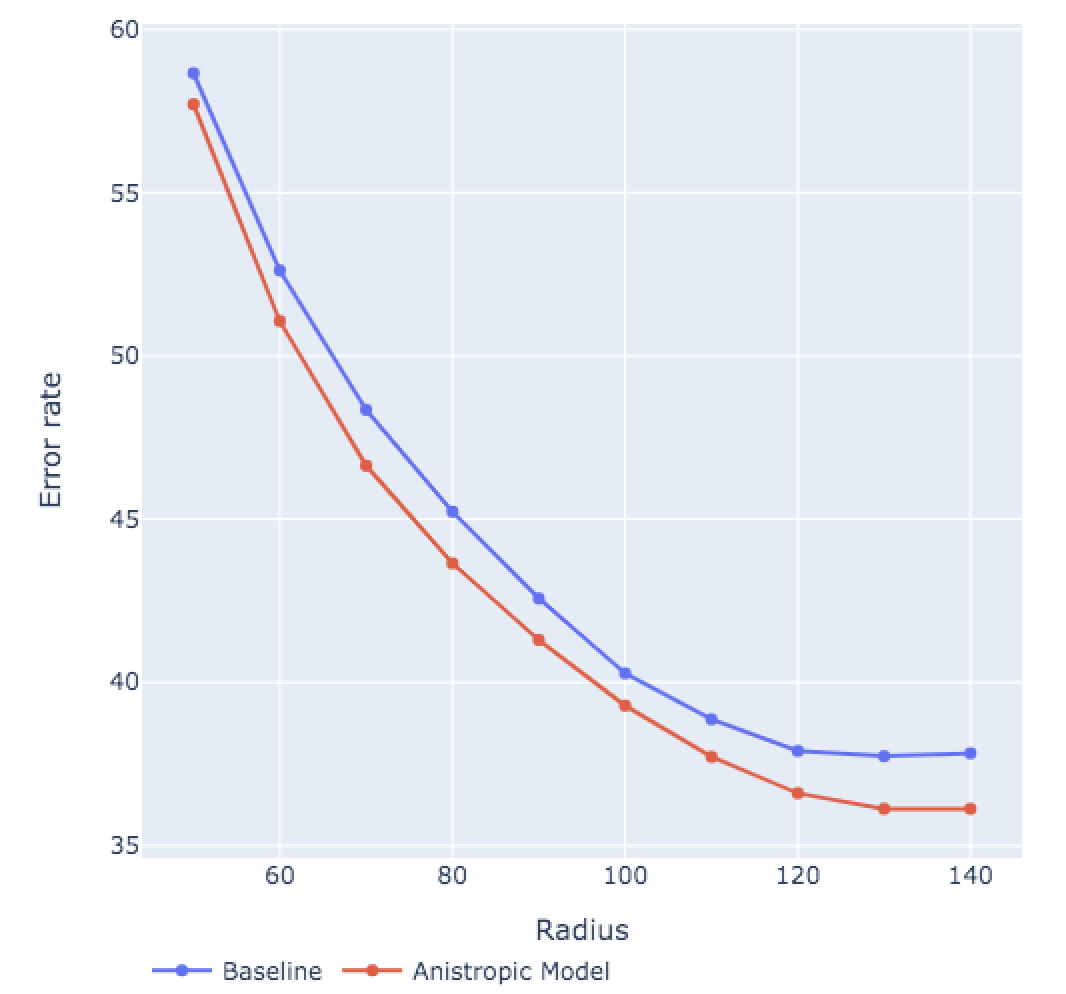}
    \caption{Results after removing high-frequency components from images. We can see that our proposed anisotropic filtering leads to models that are less reliant on high-frequency components in an image. When we remove the highest frequency components from the image(far-right), out model, outperforms baseline by close to 2.1\%. As we remove more high-frequency components, both the model suffers, but our model consistently performs better.}
    \label{fig:remove_high_frequency_component}
\end{figure}

CNNs tend to exploit the high-frequency components in an image which makes them less robust \cite{wang2020high,ilyas2019adversarial}. In this section, we show that our proposed Pix2Pix-Anisotropic model is less dependent on high-frequency components. To remove high-frequency components from an image, we first obtain the image spectrum using DFT. 
We then incrementally vary the masked region's size to retain the more/less high frequency components with $r$ denoting the mask region's side.
We can see from  Figure \ref{fig:remove_high_frequency_component} as we remove the highest frequency components, our model performs better than the baseline by +2.1\%. The same difference in normal RGB images is 0.8\%. Hence this shows that our model has less reliance on the highest frequency components.
As we remove more high-frequency components, the error increases for both models, but we are still consistently better than the baseline.

\begin{figure*}[t!]
    \centering
    \includegraphics[width=\linewidth]{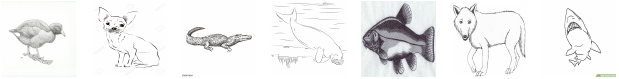}
    \caption{Examples of images from Sketch-ImageNet. Images have very little or no texture, which implies texture will have little to no impact on object classification.}
    \label{fig:sketch_images}
\end{figure*}

\begin{figure}
\includegraphics[width=\linewidth]{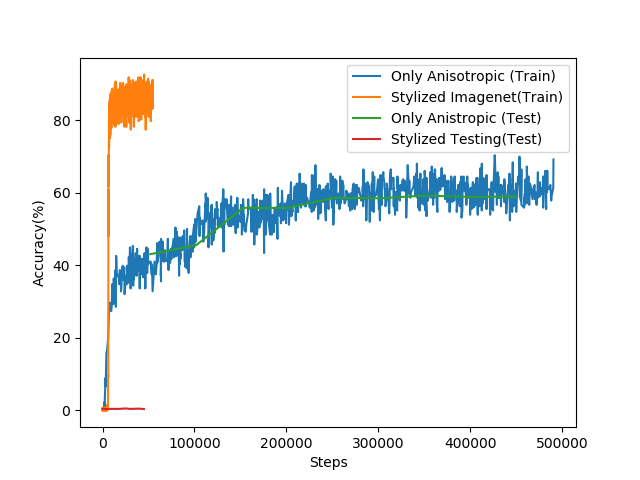}
\caption{Training plots for the both Stylized ImageNet and Standard ImageNet. The plot shows that the model trained on Stylized ImageNet quickly overfits by finding shortcuts after around 6000 steps. Therefore, it gives poor performance on downstream tasks by relying on texture based shortcuts.
}
\begin{table}
\begin{center}
\caption{Experiments discussing the confidence and entropy of Anistropic ImageNet and Standard ImageNet}
\begin{tabular}{ccc} 
\toprule
Method    & Entropy & Mean Highest probability \\
\midrule
Anistropic ImageNet     & 0.81 & 0.93   \\
Standard ImageNet   & 1.88 & 0.59   \\
\bottomrule
\end{tabular}
\label{tab: Confidence_Experiments}
\end{center}
\end{table}

\label{fig:logs_stylized}
\end{figure}

\begin{table*}[t!]
\footnotesize
\setlength\tabcolsep{0.1pt}

\begin{center}
\caption{We show additional experiments on dataset ImageNet-C\cite{Hendrycks2019BenchmarkingNN}, which evaluates model robustness to common corruptions. We can see that by focussing less on texture our model is consistently more robust than the baseline Resnet model pre-trained on ImageNet data.}
\label{tab:ImageNet_C}
\begin{tabular}{@{}l c |c c c c | c c c c | c c c  c | c c c c@{}}
\multicolumn{3}{c}{} & \multicolumn{3}{c}{Noise} & \multicolumn{4}{c}{Blur} & \multicolumn{4}{c}{Weather} & \multicolumn{4}{c}{Digital} \\
\cline{1-18}
Network  & \multicolumn{1}{c|}{\,\textbf{mCE}\,} & \scriptsize{Gauss.}
    & \scriptsize{Shot} & \scriptsize{Impulse} & \scriptsize{Defocus} & \scriptsize{Glass} & \scriptsize{Motion} & \scriptsize{Zoom} & \scriptsize{Snow} & \scriptsize{Frost} & \scriptsize{Fog} & \scriptsize{Bright} & \scriptsize{Contrast} & \scriptsize{Elastic} & \scriptsize{Pixel} & \scriptsize{JPEG}\\ \hline 
Baseline(ImageNet)    & 76.7 & 80 &	82 & 83 & 75 & 89 & 78&	80&	78&	75&	66&	57&	71&	85&	77 & 77\\
Anistropic(Ours)  & 74.85 &	72.70 &	76.65 &	78.27 & 73.73 &	87.97	& 75.99&	79.70 & 80.33  &	78.02  &	68.05  &	58.30  &	71.34  &	82.64  &	70.88  & 68 \\
\Xhline{2\arrayrulewidth}
\end{tabular}
\end{center}
\end{table*}
\subsection{Confidence of Models:}
In this section we compare the confidence and entropy of Anistropic Model and ImageNet model when both the models have given correct predictions. To find confidence, we generate the probability scores of correct class. After this we calculated the mean of correct probability scores on both the models.  As we can see from Table \ref{tab: Confidence_Experiments} that Anistropic ImageNet has larger mean which means that Anistropic ImageNet has better confidence as compared to Standard ImageNet.
We also calculate the entropy of output probability distribution from both the models. We can see from Table \ref{tab: Confidence_Experiments} Anistropic ImageNet has lower entropy scores as compared to Standard ImageNet.

\subsection{Training with Stylized ImageNet}
Figure \ref{fig:logs_stylized} shows the training curves for both Stylized ImageNet and Standard ImageNet. We see that the model quickly saturates when using Stylized ImageNet. This leads to a low performance on downstream tasks. Our hypothesis is that the model rapidly learns to exploit some regularities in the texture introduced by the GANs to easily solve the MoCoV2 and Jigsaw tasks. This means that the self-supervised model has the tendency to take shortcuts in the presence of regular textures. The aim of our paper has been to investigate such short-coimings and provide appropriate solutions to the issues.

\subsection{Results on model robustness}
 In this section, we show that suppressing texture not only helps in achieving better results but also makes our model more robust to common image corruptions.

\begin{figure}
\centering
\includegraphics[ width=0.8\linewidth]{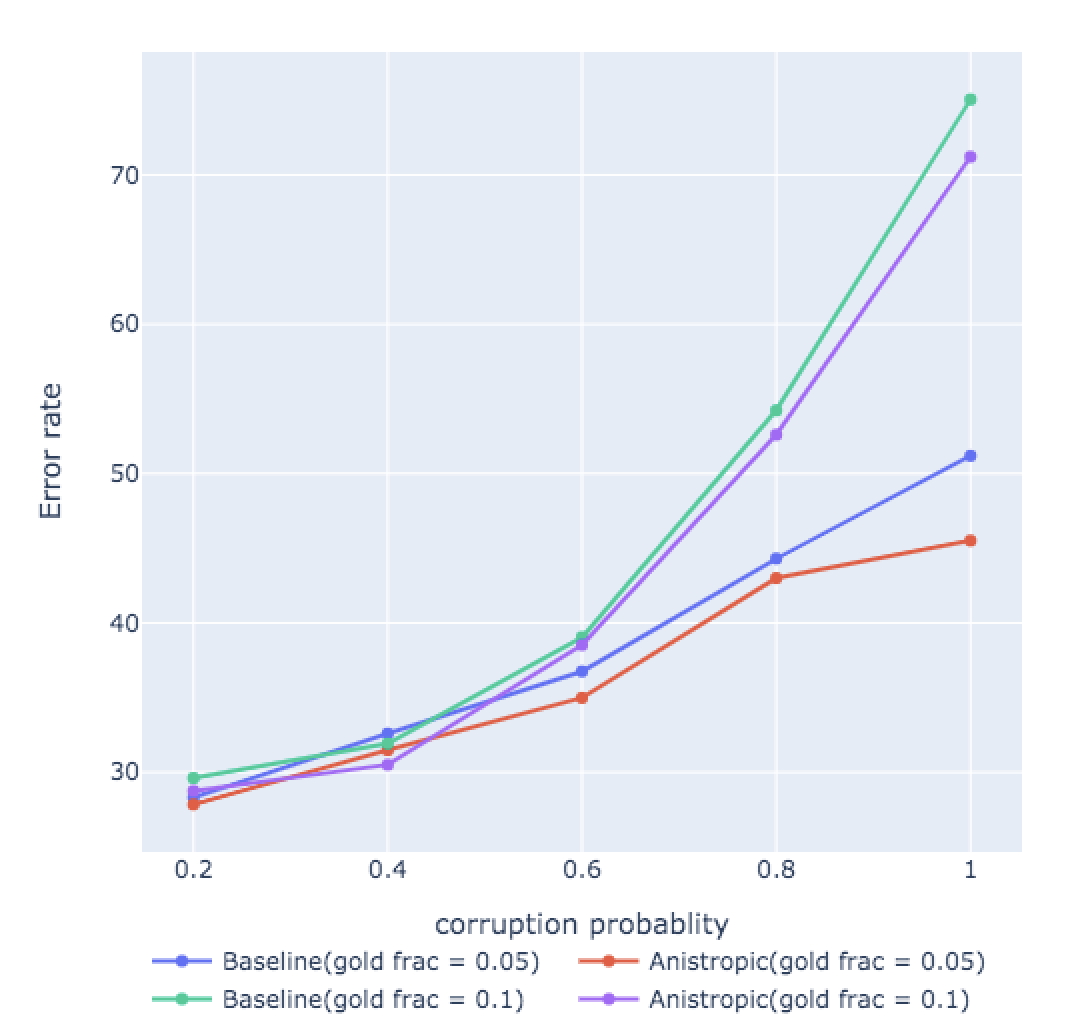}
\caption{Results on Label corruption task. The bottom two line corresponds to when the gold fraction is 0.05, and the top two lines correspond to the case when the gold fraction is 0.1.   We can see that our model consistently outperforms the baseline with a larger improvement upon increasing the corruption probability.
}
\label{fig:label_corruption}
\end{figure}

\textbf{Label Corruption Task:}
We show results on the label corruption task in Fig \ref{fig:label_corruption}. In this task, \cite{hendrycks2019selfsupervised} only some fraction(gold fraction) of data is clean, and the label of the rest of the data is corrupted with some probability. The task is to use both of these datasets for supervised classification.
 We use CIFAR100 as our dataset, and we augment the CIFAR100 dataset with Anisotropic diffused images. We create a dataset double the size of the original dataset and use it for the task of Label Corruption \cite{hendrycks2019selfsupervised}. 
We can see from the results that we have consistent improvement over the baseline. With an increase in corruption probability, our results improve, even more, implying that focusing on higher-level features helps even more in extreme corruptions.

\textbf{Results on ImageNet-C:}
We show additional experiments on dataset ImageNet-C\cite{Hendrycks2019BenchmarkingNN}, which evaluates model robustness to common corruptions. ImageNet-C[1] dataset has a total of 15 corruptions, as shown in Table \ref{tab:ImageNet_C}. 
We improve(+1.85\%) upon the baseline Resnet model pre-trained on ImageNet data and perform especially well in the cases of noise corruptions since we train on anisotropic images, which removes high-frequency noise from images.

\section{Jigsaw Task}
In addition to MoCoV2 we also show results with Jigsaw pretext task. 
\subsection{Method}

The goal in Jigsaw is to infer correct ordering of the given regions from an image. Following \cite{Noroozi2016UnsupervisedLO}, the typical setting is to divide an image into nine non-overlapping square patches and randomly shuffle them. A CNN is trained to predict the original permutation of these patches in the image. Jigsaw++ \cite {Noroozi_2018_CVPR} extended this idea and replaced some patches with random patches. 
These patch based methods come with their own issues and there has been some recent effort to solve them. \cite {Mundhenk2017ImprovementsTC} describe an easy short-cut CNNs take that utilizes the chromatic aberration produced due to different wavelengths exiting the lens at different angles. The authors provided a solution to the problem of chromatic aberration by removing cues from the images and also allowing the color pattern to be partially preserved. They also address the problem of true spatial extent that network sees in patch based methods by yoking the patch jitter to create a random crop effect.

\subsection{Results using Jigsaw Pretraining}

The baseline model in the case of Jigsaw is pre-trained using the standard ImageNet dataset. Unlike \cite{Noroozi2016UnsupervisedLO}, we use ResNet18 as our backbone instead of Alexnet\cite{Krizhevsky2017ImageNetCW} to take advantage of deeper layers and capture better image representations. 
We obtained these results by building on top of a publicly available implementation\footnote{\href{https://github.com/bbrattoli/JigsawPuzzlePytorch}{https://github.com/bbrattoli/JigsawPuzzlePytorch}}.
Table \ref{tab:Stylzied_Experiments} shows results for Jigsaw models trained and tested on different datasets. We observe that the Jigsaw model trained on the Cartoon dataset outperforms the baseline methods by 2.52 mAP and Anisotropic ImageNet outperforms the baseline methods by 1.8 mAP on the PASCAL VOC image classification dataset.
On object detection Bilateral ImageNet outperforms the baseline Jigsaw model by 0.78 mAP. 
On semantic segmentation Anisotropic Imagenet outperforms the baseline Jigsaw models by 8.1 mAP. Traditionally semantic segmentation has been a difficult task for Self-Supervised methods \cite{Noroozi2016UnsupervisedLO,Caron2018DeepCF} and improvement of this order on semantic segmentation shows the effectiveness of removing texture.

We also show results on ImageNet classification as the downstream task. 
Due to its
large scale, it is usually infeasible to fine-tune the whole network for the final task every time.
Therefore, following prior work \cite{Caron2018DeepCF}, we only fine-tune a linear classifier. 
The inputs to this classifier are the features from a convolution layer in the network. Note that while fine-tuning for the final task, we keep the backbone frozen. Therefore, the performance of the linear classifier can
be seen as a direct representation of the quality of the features obtained from the CNN.  
We report
the results of this experiment on ImageNet in Table \ref{ref:Jigsaw_Imgagenet_Experiments}. 
Adding the Anisotropic ImageNet dataset to this model gives a further improvement of 0.5\%.

\subsection{Jigsaw using Alexnet as backbone}
Our improvement when using Anisotropic ImageNet is not restricted to the backbone. Traditionally in Self-Supverised learning one of the most followed architectures is Alexnet \cite{Noroozi2016UnsupervisedLO, Doersch2015UnsupervisedVR, Caron2018DeepCF}. Following these methods, we also show results on Alexnet backbone.  
In Table. \ref{tab: Alexnet_Experiments} we show results on VOC Classification when using Alexnet as the backbone. We obtain an improvement of 0.67 mAP over the baseline.

\subsection{Patch-wise Anisotropic Diffusion} 
In our best performing model, we considered all the patches for the jigsaw task to either come from the standard ImageNet or anisotropic diffusion filtered ImageNet. What if each of the 9 patches for the Jigsaw task could be either a standard patch or filtered patch?  For this experiment we randomly choose a patch from the standard dataset or the filtered dataset, with equal probability. This is a much more extreme form of data augmentation and considerably increases the difficulty of the task. We got an improvement of 0.6 mAP over the baseline model for the classification task. However this is 1.1 mAP lower then the doing Anistropic Diffusion on whole image.

\begin{table*}
\begin{center}
\caption{Comparison of our approach with Jigsaw baseline methods. Using our best model, we improve 2.52 mAP in VOC classification , 0.78 mAP on VOC detection and 8.1 mAP on VOC semantic segmentation(SS) over the baseline models. Note that Stylized ImageNet performs poorly on VOC classification due to the visual shortcuts.}
\begin{tabular}{ccccc}  
\toprule
Method    & Dataset Size & VOC Cls. & VOC Det. & SS \\
\midrule
Baseline     & 1.2M & 74.82    & 61.98  & 27.1  \\
Stylized \cite{Geirhos2018ImageNettrainedCA}     & 1.2M & 13.81    & 28.13  &10.12   \\
Gaussian ImageNet & 2$\times$1.2M    & 75.49    & 62.39    &27.9  \\
Bilateral ImageNet & 2$\times$1.2M    & 74.55    & \textbf{62.74}    &28.9   \\
Only Anisotropic   & 1.2M & 74.52    & {61.85}  & 32.7     \\
Anisotropic ImageNet & 2$\times$1.2M    & {76.77}    & 61.59   &\textbf{35.2}    \\
Cartoon ImageNet & 2$\times$1.2M    & \textbf{77.34}    & 59.31   &  34.1\\
\bottomrule
\end{tabular}
\label{tab:Stylzied_Experiments}
\end{center}
\end{table*}
\begin{table*}[t!]
\caption{ImageNet classification by finetuning the last FC layer. Features from the conv layers are kept unchanged. This experiment helps evaluate the quality of features learnt by the convolutional layers.}
\vspace{-0.5em}
\begin{center}
{
\begin{tabular}{cccccc}  
\toprule
Method    & Dataset Size & VOC Cls & VOC Det. & ImageNet Cls. Acc \\
\midrule
Jigsaw Baseline     & 1.2M & 74.82    & 61.98  &26.17  \\
Jigsaw anisotropic   & 2$\times$1.2M& 76.77    & 61.59  & 26.67    \\
\bottomrule
\end{tabular}
\label{ref:Jigsaw_Imgagenet_Experiments}
}
\vspace{-1em}
\end{center}
\end{table*}
\begin{table*}
\begin{center}
\caption{Experiments with Alexnet as the backbone. Ideas of anisotropic diffusion filter can extend to other architectures like Alexnet. The Anistropic ImageNet model improves over the baseline by 0.67 mAP}
\begin{tabular}{cc} 
\toprule
Method    & VOC 2007 Classification \\
\midrule
Jigsaw Baseline(Our Implementation)     & 65.21    \\
Jigsaw anisotropic   & 65.88    \\
\bottomrule
\end{tabular}
\label{tab: Alexnet_Experiments}
\end{center}
\end{table*}



\section{Anistropic Images}
\begin{figure*}%
    \centering
        \centering
        \includegraphics[width==0.35\linewidth]{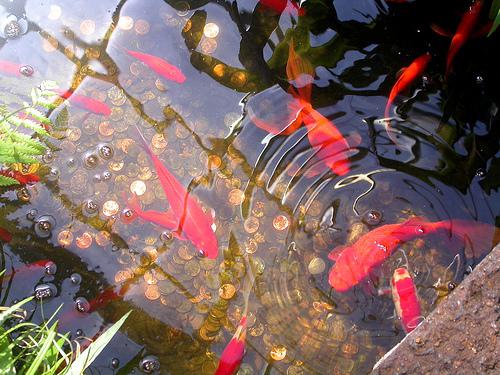}
        \includegraphics[width=0.35\linewidth]{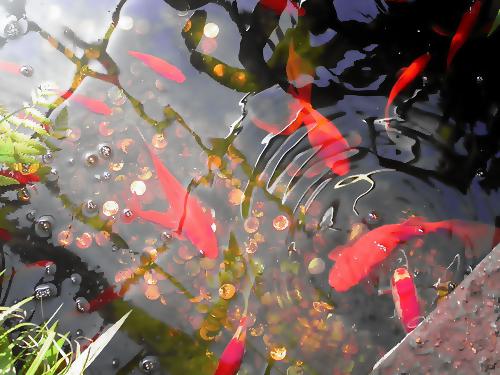}
        \centering
        \includegraphics[width=0.35\linewidth]{images/n01443537_24724_og.jpg}
    \\
        \centering
        \includegraphics[width=0.35\linewidth]{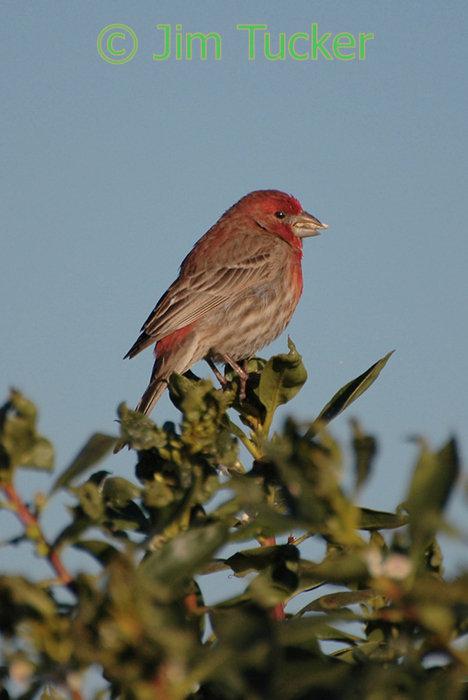}
        \centering
        \includegraphics[width=0.35\linewidth]{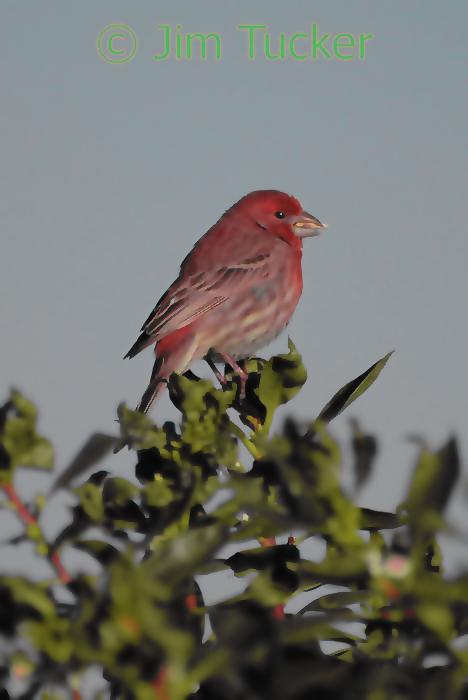}
    \\
        \centering
        \includegraphics[width=0.35\linewidth,height=7cm]{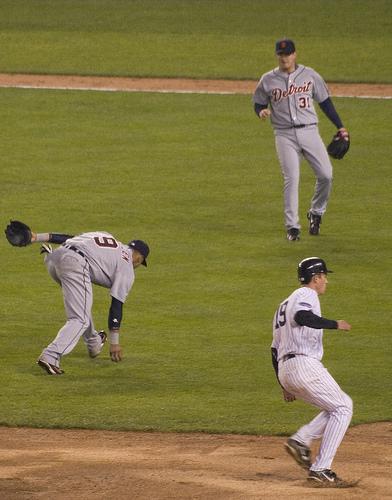}
        \centering
        \includegraphics[width=0.35\linewidth,height=7cm]{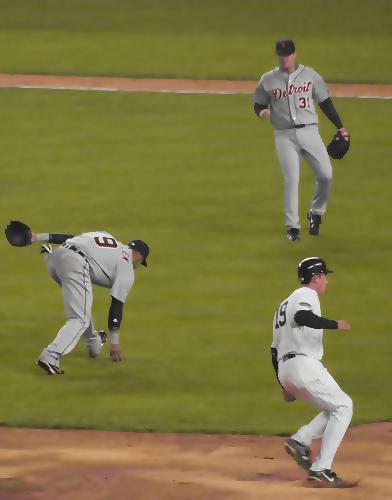}
    \caption{Original images (left)and images obtained after anisotropic diffusion (right). Most of the texture information in the images has been smoothed out by the filter while retaining the shape information. This forces the network to capture higher-level semantics without relying on low-level texture cues}
    \label{fig:example1}
    \end{figure*}

\begin{figure*}%
 \centering
        \centering
        \includegraphics[width=0.35\linewidth]{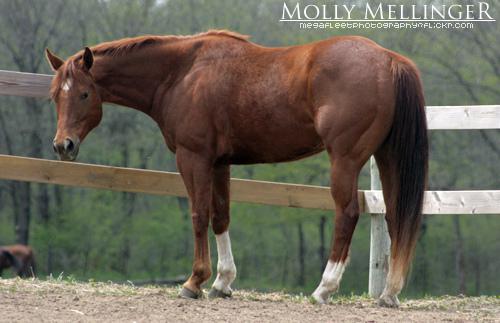}
        \centering
        \includegraphics[width=0.35\linewidth]{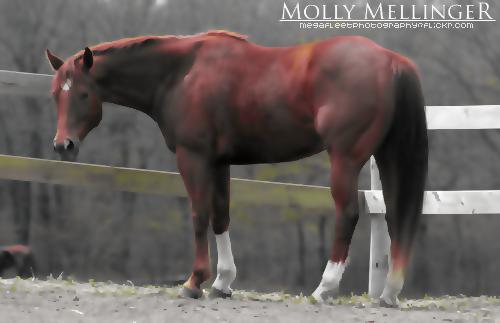}
    \\
        \centering
        \includegraphics[width=0.35\linewidth]{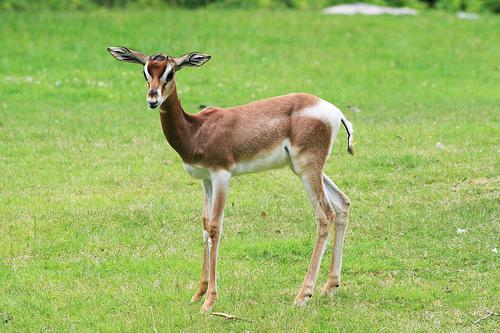}
        \centering
        \includegraphics[width=0.35\linewidth]{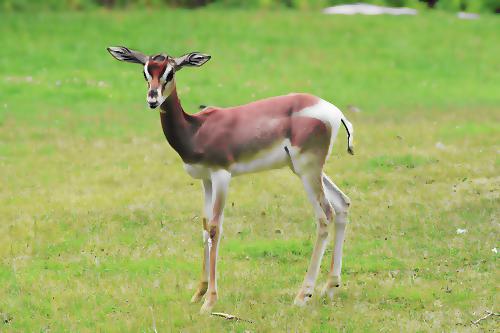}
    \\
        \centering
        \includegraphics[width=0.35\linewidth]{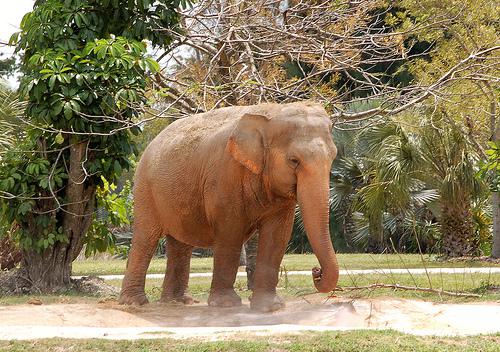}
        \centering
        \includegraphics[width=0.35\linewidth]{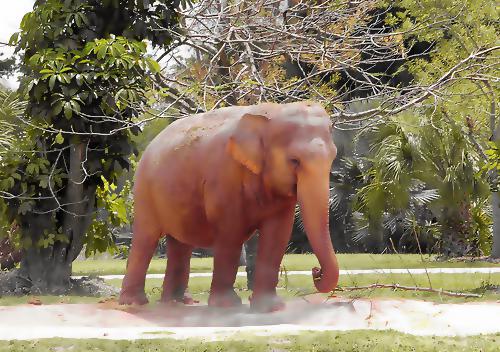}
    \\
        \centering
        \includegraphics[width=0.35\linewidth]{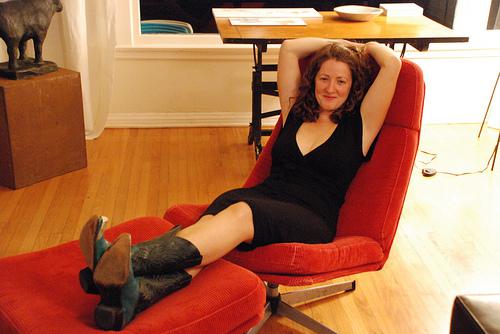}
        \centering
        \includegraphics[width=0.35\linewidth]{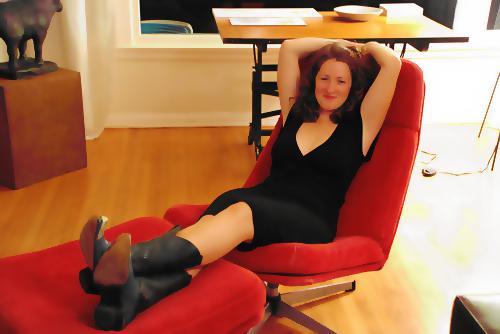}
    \\
    \caption{Original images (left)and images obtained after anisotropic diffusion (right). Most of the texture information in the images has been smoothed out by the filter while retaining the shape information. This forces the network to capture higher-level semantics without relying on low-level texture cues}
    \label{fig:example2}%
\end{figure*}

We show some more examples of Anistropic images obtained by applying Anisotropic diffusion filiters to images from ImageNet \cite{imagenet_cvpr09} in figures \ref{fig:example1} and \ref{fig:example2}.  Notice how the images lose texture information. This makes it more difficult for models to find shortcuts. This, in turn, leads to better semantic representations learned by the model which leads to higher performance on downstream tasks.


\section{Saliency Maps}
\subsection{Sketch-ImageNet Saliency Maps}
\begin{figure*}[h!]
\centering
\begin{tabular}{cccccc}
\rotatebox{90}{Anisotropic}&
\includegraphics[width=0.15\linewidth]{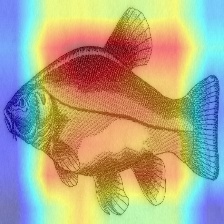}&
\includegraphics[width=0.15\linewidth]{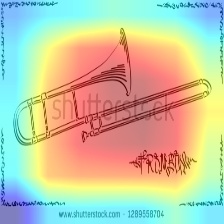}&
\includegraphics[width=0.15\linewidth]{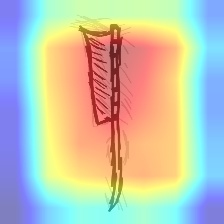}&
\includegraphics[width=0.15\linewidth]{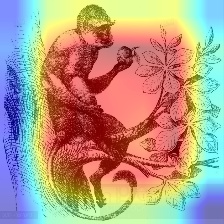}&
\includegraphics[width=0.15\linewidth]{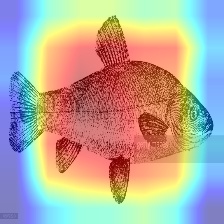}\\
\rotatebox{90}{ImageNet}&
\includegraphics[width=0.15\linewidth]{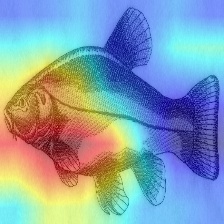}&
\includegraphics[width=0.15\linewidth]{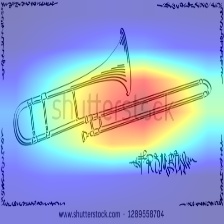}&
\includegraphics[width=0.15\linewidth]{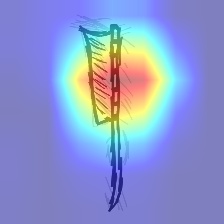}&
\includegraphics[width=0.15\linewidth]{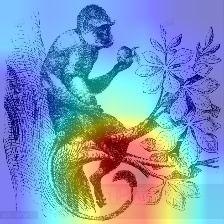}&
\includegraphics[width=0.15\linewidth]{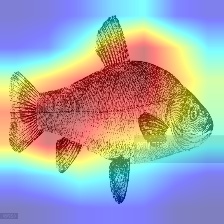}\\
\\
&(a)&(b)&(c)&(d)&(e)\\
\end{tabular}
  \caption{Saliency maps on few randomly selected images from Sketch-ImageNet. We can see from saliency maps that Anisotropic model has bigger saliency maps which might be the reason for the correct prediction.}
    \label{fig:sketch_saliency}

\end{figure*}

In Fig \ref{fig:sketch_saliency} we show some of saliency maps for Sketch-ImageNet images. We can see from saliency maps that Anistropic ImageNet has broader saliency map and has better coverage of the object as compared to ImageNet model. 
\subsection{Saliency Maps for Anistropic ImageNet and Standard ImageNet models}
\begin{figure*}[h!]
\centering
\begin{tabular}{cccccc}
\rotatebox{90}{Anisotropic}&
\includegraphics[width=0.15\linewidth]{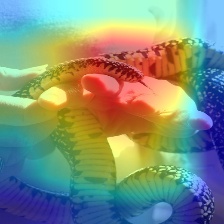}&
\includegraphics[width=0.15\linewidth]{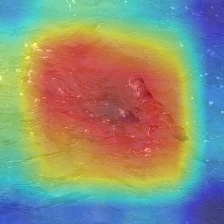}&
\includegraphics[width=0.15\linewidth]{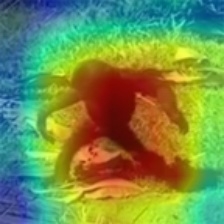}&
\includegraphics[width=0.15\linewidth]{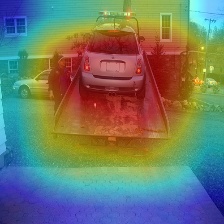}&
\includegraphics[width=0.15\linewidth]{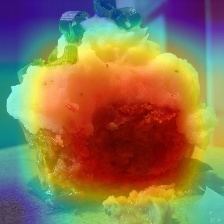}\\
\rotatebox{90}{ImageNet}&
\includegraphics[width=0.15\linewidth]{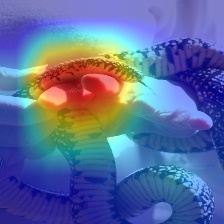}&
\includegraphics[width=0.15\linewidth]{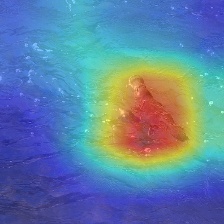}&
\includegraphics[width=0.15\linewidth]{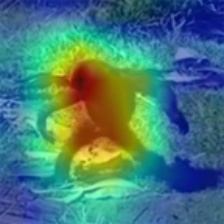}&
\includegraphics[width=0.15\linewidth]{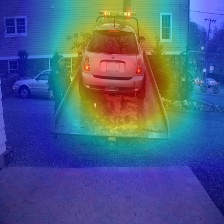}&
\includegraphics[width=0.15\linewidth]{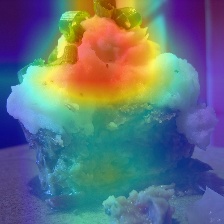}\\
\\
&(a)&(b)&(c)&(d)&(e)\\
\end{tabular}
  \caption{Saliency maps when Anistropic Model had correct predictions and ImageNet model has wrong predictions.}
    \label{fig:anis_correct}

\end{figure*}

\begin{figure*}[h!]
\centering
\begin{tabular}{cccccc}
\rotatebox{90}{Anisotropic}&
\includegraphics[width=0.15\linewidth]{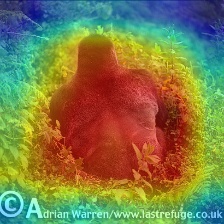}&
\includegraphics[width=0.15\linewidth]{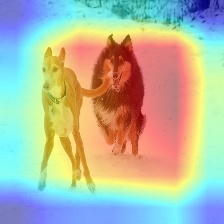}&
\includegraphics[width=0.15\linewidth]{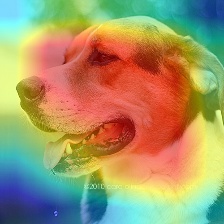}&
\includegraphics[width=0.15\linewidth]{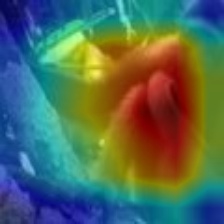}&
\includegraphics[width=0.15\linewidth]{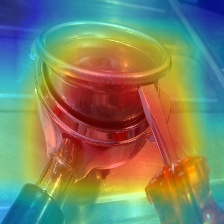}\\
\rotatebox{90}{ImageNet}&
\includegraphics[width=0.15\linewidth]{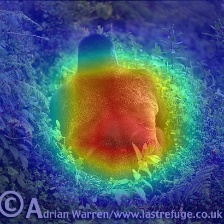}&
\includegraphics[width=0.15\linewidth]{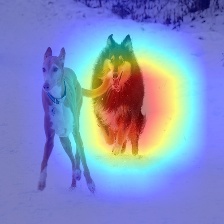}&
\includegraphics[width=0.15\linewidth]{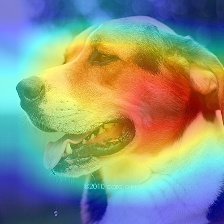}&
\includegraphics[width=0.15\linewidth]{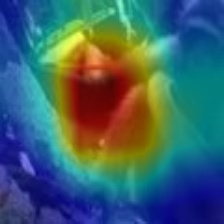}&
\includegraphics[width=0.15\linewidth]{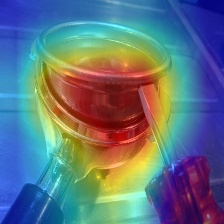}\\
\\
&(a)&(b)&(c)&(d)&(e)\\
\end{tabular}
  \caption{Saliency maps when both model have wrong predictions.}
    \label{fig:both_wrong}

\end{figure*}

\begin{figure*}[h!]
\centering
\begin{tabular}{cccccc}

\rotatebox{90}{Anisotropic}&
\includegraphics[width=0.15\linewidth]{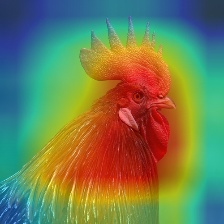}&
\includegraphics[width=0.15\linewidth]{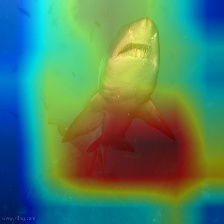}&
\includegraphics[width=0.15\linewidth]{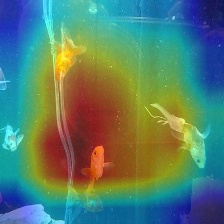}&
\includegraphics[width=0.15\linewidth]{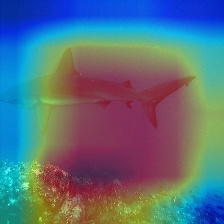}&
\includegraphics[width=0.15\linewidth]{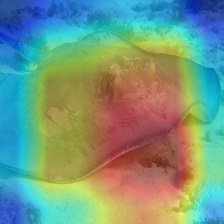}\\
\rotatebox{90}{ImageNet}&
\includegraphics[width=0.15\linewidth]{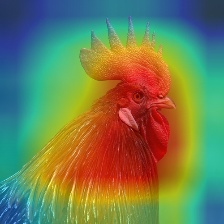}&
\includegraphics[width=0.15\linewidth]{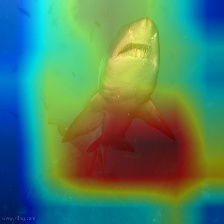}&
\includegraphics[width=0.15\linewidth]{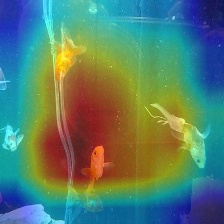}&
\includegraphics[width=0.15\linewidth]{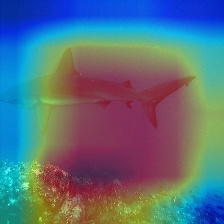}&
\includegraphics[width=0.15\linewidth]{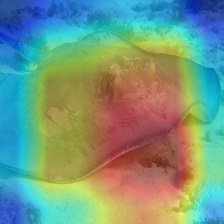}\\
\\
&(a)&(b)&(c)&(d)&(e)\\
\end{tabular}
  \caption{Saliency maps when both model have correct predictions.}
    \label{fig:both_correct}

\end{figure*}

\begin{figure*}[h!]
\centering
\begin{tabular}{cccccc}

\rotatebox{90}{Anisotropic}&
\includegraphics[width=0.15\linewidth]{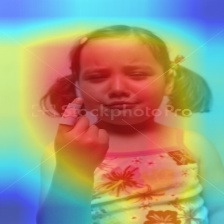}&
\includegraphics[width=0.15\linewidth]{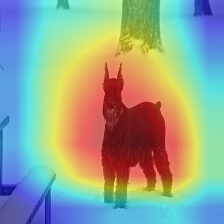}&
\includegraphics[width=0.15\linewidth]{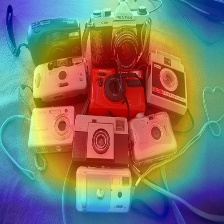}&
\includegraphics[width=0.15\linewidth]{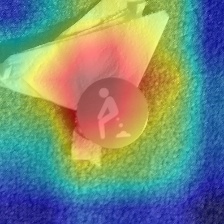}&
\includegraphics[width=0.15\linewidth]{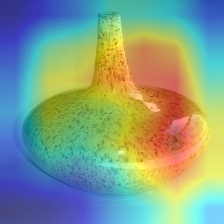}\\
\rotatebox{90}{ImageNet}&
\includegraphics[width=0.15\linewidth]{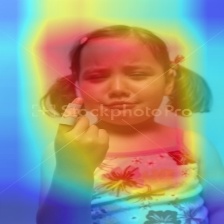}&
\includegraphics[width=0.15\linewidth]{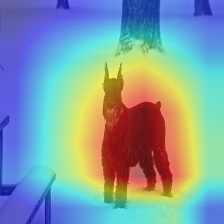}&
\includegraphics[width=0.15\linewidth]{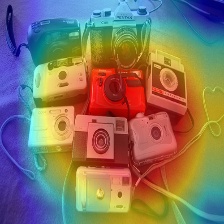}&
\includegraphics[width=0.15\linewidth]{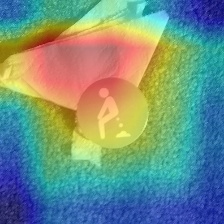}&
\includegraphics[width=0.15\linewidth]{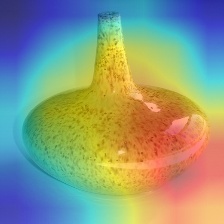}\\
\\
&(a)&(b)&(c)&(d)&(e)\\
\end{tabular}
  \caption{Saliency maps when ImageNet model has correct predictions and Anistropic model has wrong predictions.}
    \label{fig:img_correct}

\end{figure*}

We also show some addtional saliency maps in Figure \ref{fig:anis_correct}, Figure \ref{fig:both_correct}, Figure \ref{fig:both_wrong} and Figure \ref{fig:img_correct} corresponding to both the models. We can see from the figures that Anistropic ImageNet has in general diffused saliency maps.

\begin{table*}[t!]
\begin{center}
\caption{Results on Label corruption task. We can see that our model consistently outperforms the baseline with larger improvement upon increasing the corruption probability.}
\begin{tabular}{cccccc}  
\toprule
Corruption probability    & Gold Fraction & Baseline(Error) & Anistropic Model(Error) \\
\midrule
0.2 &	0.05 &	29.61 &\textbf{	28.75 } \\
0.2 &	0.1 &	28.31 &	\textbf{27.84} \\
0.4 &	0.05 &	31.92 &	\textbf{30.5} \\
0.4 &	0.1 &	32.59 &	\textbf{31.48} \\
0.6 &	0.05 &	39.04 &	\textbf{38.52} \\
0.6 &	0.1 &	36.75 &	\textbf{34.98} \\
0.8 &	0.05 &	54.23 &	\textbf{52.6} \\
0.8 &	0.1 &	44.31 &	\textbf{43} \\
1.0 &	0.05 &	75.05 &	\textbf{71.21} \\
1.0 &	0.1 &	51.19 &	\textbf{45.51} \\
\bottomrule
\end{tabular}
\label{tab:Label_Corruption}
\end{center}
\end{table*}

\section{Implementation details}
\paragraph{Training Details.}
With image classification as the downstream task, we train our network for 90,000 iterations with an
initial learning rate of 0.003 following \cite{Caron2018DeepCF}. 

For object detection we report our
results for Faster-RCNN \cite{Ren2015FasterRT} using our pre-trained model as backbone. 
We tune hyper-parameters using the validation set.  
For object detection, we follow the details of \cite{Ren2015FasterRT} to train a model; 10 epochs with an initial learning rate of 0.001.

For semantic segmentation, we report our results on FCN\cite{Shelhamer_2017} using our pretrained model as backbone. 
We train the FCN model for 30 epochs using an initial learning rate of 0.01.

\noindent\textbf{ImageNet.} 
We use ImageNet for all training and evaluation of image classification accuracy. 
For self-supervised learning, we follow \cite{Caron2018DeepCF,he2019momentum}; we train linear classifiers using features obtained from the final Residual block by freezing all convolutional layers. 
The performance of these linear classifiers is meant to evaluate the quality of the feature
representations learnt by the convolutional layers, since the backbone is completely frozen and only
fully-connected layers are being trained. 
We chose hyper-parameters using the validation set and report performance on the ImageNet validation set.

Note that since we use ImageNet to pre-train for self-supervised learning, there is no domain difference when we conduct inference on ImageNet, but with VOC there is.  
With the VOC results, we validate that the gain by our method is particularly large when there is domain shift.\\
\noindent\textbf{MoCo-v2 task.} For MoCo-v2 we use  higher lr of 0.3 as it gives better performance for our task.

\noindent\textbf{Jigsaw task.} 
In Jigsaw \cite{Noroozi2016UnsupervisedLO} the image is divided into 9 non-overlapping square patches. 
We select 1,000 from the 9! possible permutations. All of our primary experiments on Jigsaw use ResNet18 as the backbone \cite{He2015}. We train the Jigsaw task for 90 epochs, with an initial learning rate of 0.01. The learning rate is reduced by a factor of 0.1 after $(30,30,20,10)$ epochs. We use the same data augmentation as in \cite{Noroozi2016UnsupervisedLO}. In MoCo \cite{He2015}, we use ResNet50\cite{He2015} as the backbone, following the same procedure as mentioned in \cite{he2019momentum}.

\noindent\textbf{PASCAL VOC.} 
Following \cite{Caron2018DeepCF} and \cite{chen2020improved}, 
we evaluate image classification and object detection on the PASCAL VOC dataset \cite{Everingham2009ThePV}. 
It contains about 5,000 images in the train-val set belonging to 20 classes. 
Note that the image classification task is multi-label. Therefore, the metric used for evaluating both image classification and object detection is the mean Average Precision (mAP). 
\paragraph{Training Details for Object detection for COCO based metrics:}
We report object detection results on \cite{Ren2015FasterRT} C4 backbone which is finetuned end to end on VOC07+12 trainval dataset and evaluated on the VOC 07
test set using the COCO suite of metrics. 

\noindent\textbf{Other Details.}
We use 4 Nvidia GTX 1080 Ti for all experiments.
Pretraining on Jigsaw takes 3 days on the standard ImageNet dataset. The SGD optimizer with momentum was used for all experiments with momentum of 0.9 and weight decay of $5\times10^{-4}$. Cross-entropy loss was used for all experiments, mini-batch size was set to 256. Pretraining on MoCoV2 takes 6 days on 4 Nvidia P100 machines. We set all other hyperparamters following \cite{chen2020improved}.

\begin{table*}[t!]
\centering
\caption{Comparison between Stylized ImageNet and our Anisotropic ImageNet. Following \cite{Geirhos2018ImageNettrainedCA}, we use ResNet50 as our backbone. We finetune our models on only the ImageNet dataset. We can see that on ImageNet classification and object detection, Anisotropic ImageNet and Stylized ImageNet have very similar performance.}
\label{tab: Stylized_Classification_experiments}
\begin{tabular}{ccccc}  
\toprule
Method    & Finetune & Top-1 Accuracy & Top-5 Accuracy & OBJ Detection \\
\midrule
Stylized Imagenet     & - & 74.59    & 92.14 &  70.6   \\
Stylized Imagenet    & IN & 76.72    & 93.28 &  75.1  \\
Anisotropic Imagenet   & - & 68.38    & 87.19 & -\\
Anisotropic Imagenet   & IN & 76.71    & 93.26 &   74.27 \\

Cartoon Imagenet   & IN & 76.22    & 93.12 & 72.31   \\
\bottomrule
\end{tabular}
\end{table*}



\section{Label Corruption Task}
We show the full results of the label corruption task in Table \ref{tab:Label_Corruption}.

\begin{table*}
\begin{center}
\caption{Experiments discussing the confidence and entropy of Anistropic ImageNet and Standard ImageNet}
\begin{tabular}{ccc} 
\toprule
Method    & Entropy & Mean Highest probability \\
\midrule
Anistropic ImageNet     & 0.81 & 0.93   \\
Standard ImageNet   & 1.88 & 0.59   \\
\bottomrule
\end{tabular}
\label{tab: Confidence_Experiments}
\end{center}
\end{table*}

\begin{figure*}[t!]
\centering
\resizebox{1\linewidth}{!}{
\begin{tabular}{ccccccc}
{\rotatebox{90}{\parbox{0.15\linewidth}{\centering Anisotropic}}}&
{\rotatebox{90}{\parbox{0.15\linewidth}{\centering Correct}}}&
\includegraphics[width=0.18\linewidth]{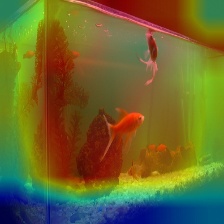}&
\includegraphics[width=0.18\linewidth]{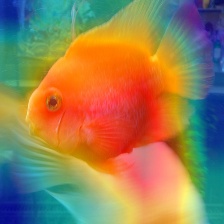}&
\includegraphics[width=0.18\linewidth]{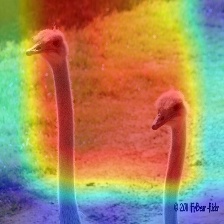}&
\includegraphics[width=0.18\linewidth]{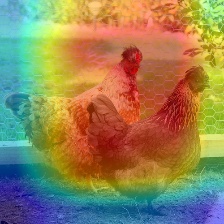}&
\includegraphics[width=0.18\linewidth]{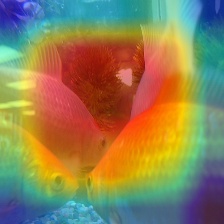}\\
\rotatebox{90}{\parbox{0.15\linewidth}{\centering ImageNet}}&
\rotatebox{90}{\parbox{0.15\linewidth}{\centering Correct}}&
\includegraphics[width=0.18\linewidth]{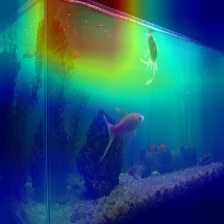}&
\includegraphics[width=0.18\linewidth]{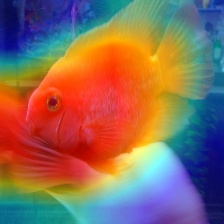}&
\includegraphics[width=0.18\linewidth]{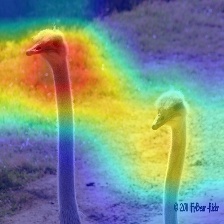}&
\includegraphics[width=0.18\linewidth]{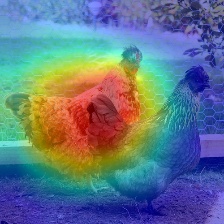}&
\includegraphics[width=0.18\linewidth]{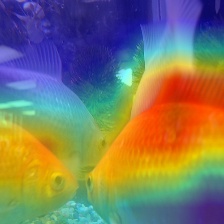}
\\
&&(a)&(b)&(c)&(d)&(e)\\
\hline
\\
\\
{\rotatebox{90}{\parbox{0.15\linewidth}{\centering Anisotropic}}}&
{\rotatebox{90}{\parbox{0.15\linewidth}{\centering Correct}}}&
\includegraphics[width=0.18\linewidth]{images/saliency_maps/anis_correct_anis_model/ILSVRC2012_val_00000061.jpeg}&
\includegraphics[width=0.18\linewidth]{images/saliency_maps/anis_correct_anis_model/ILSVRC2012_val_00000114.jpeg}&
\includegraphics[width=0.18\linewidth]{images/saliency_maps/anis_correct_anis_model/ILSVRC2012_val_00000293.jpeg}&
\includegraphics[width=0.18\linewidth]{images/saliency_maps/anis_correct_anis_model/ILSVRC2012_val_00000317.jpeg}&
\includegraphics[width=0.18\linewidth]{images/saliency_maps/anis_correct_anis_model/ILSVRC2012_val_00000329.jpeg}\\
{\rotatebox{90}{\parbox{0.15\linewidth}{\centering ImageNet}}}&
{\rotatebox{90}{\parbox{0.15\linewidth}{\centering Incorrect}}}&
\includegraphics[width=0.18\linewidth]{images/saliency_maps/anis_correct_img_model/ILSVRC2012_val_00000061.jpeg}&
\includegraphics[width=0.18\linewidth]{images/saliency_maps/anis_correct_img_model/ILSVRC2012_val_00000114.jpeg}&
\includegraphics[width=0.18\linewidth]{images/saliency_maps/anis_correct_img_model/ILSVRC2012_val_00000293.jpeg}&
\includegraphics[width=0.18\linewidth]{images/saliency_maps/anis_correct_img_model/ILSVRC2012_val_00000317.jpeg}&
\includegraphics[width=0.18\linewidth]{images/saliency_maps/anis_correct_img_model/ILSVRC2012_val_00000329.jpeg}
\\
&&(f)&(g)&(h)&(i)&(j)\\
\hline
\\
\\

{\rotatebox{90}{\parbox{0.15\linewidth}{\centering Anisotropic}}}&
{\rotatebox{90}{\parbox{0.15\linewidth}{\centering Incorrect}}}&
\includegraphics[width=0.18\linewidth]{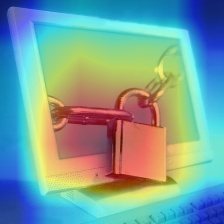}&
\includegraphics[width=0.18\linewidth]{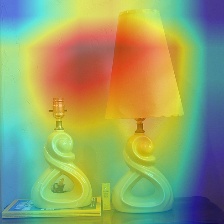}&
\includegraphics[width=0.18\linewidth]{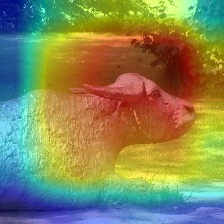}&
\includegraphics[width=0.18\linewidth]{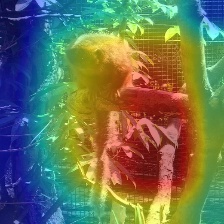}&
\includegraphics[width=0.18\linewidth]{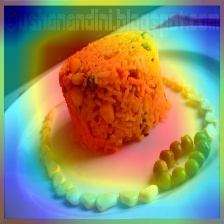}\\
{\rotatebox{90}{\parbox{0.15\linewidth}{\centering ImageNet}}}&
{\rotatebox{90}{\parbox{0.15\linewidth}{\centering Correct}}}&
\includegraphics[width=0.18\linewidth]{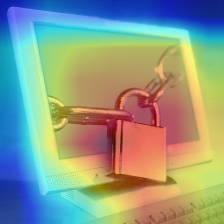}&
\includegraphics[width=0.18\linewidth]{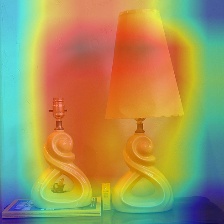}&
\includegraphics[width=0.18\linewidth]{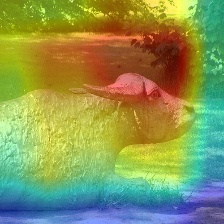}&
\includegraphics[width=0.18\linewidth]{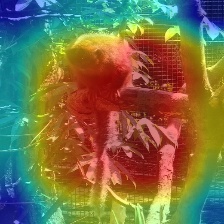}&
\includegraphics[width=0.18\linewidth]{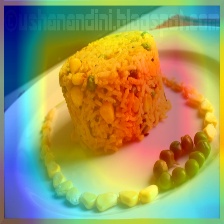}
\\ 
&&(k)&(l)&(m)&(n)&(o)\\
\end{tabular}
}
\caption{Saliency maps on three different set of images. The text on the left of the row indicates whether Anisotropic model or ImageNet model was used. The first two rows show the saliency maps where both model gave correct predictions. We can see from saliency maps that the Anisotropic model has more diffused saliency maps. The second two rows show the saliency maps where Anisotropic model gave correct predictions and ImageNet model gave wrong predictions. The failure of ImageNet model might be due to it not attending to whole object. The last two rows show the saliency maps where Anisotropic model gives incorrect predictions and ImageNet model gives correct predictions. Even in this failure mode, the Anisotropic model gives diffused saliency maps.}
  \label{fig:correct_incorrect}
\end{figure*}

\begin{figure*}[h!]
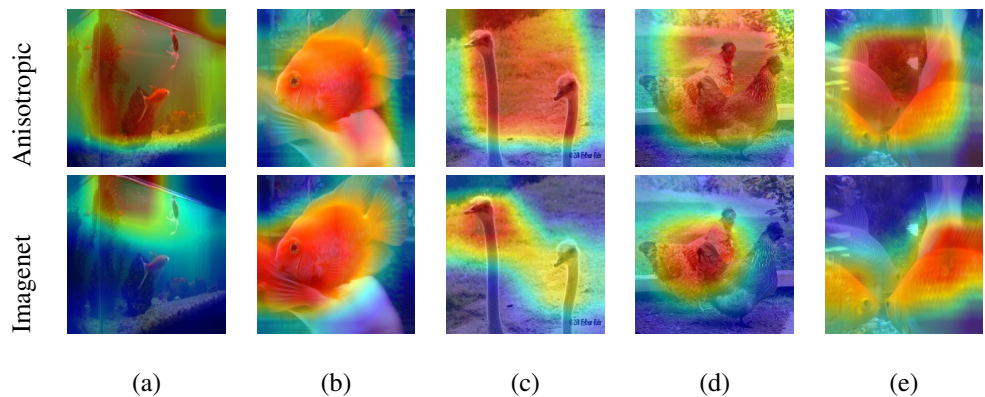

\centering
\begin{tabular}{cccccc}
\rotatebox{90}{Anisotropic}&
\includegraphics[width=0.15\linewidth]{images/saliency_maps/both_correct_anis_model/ILSVRC2012_val_00000262.jpeg}&
\includegraphics[width=0.15\linewidth]{images/saliency_maps/both_correct_anis_model/ILSVRC2012_val_00000994.jpeg}&
\includegraphics[width=0.15\linewidth]{images/saliency_maps/both_correct_anis_model/ILSVRC2012_val_00001031.jpeg}&
\includegraphics[width=0.15\linewidth]{images/saliency_maps/both_correct_anis_model/ILSVRC2012_val_00001114.jpeg}&
\includegraphics[width=0.15\linewidth]{images/saliency_maps/both_correct_anis_model/ILSVRC2012_val_00000307.jpeg}\\
\rotatebox{90}{Imagenet}&
\includegraphics[width=0.15\linewidth]{images/saliency_maps/both_correct_img_model/ILSVRC2012_val_00000262.jpeg}&
\includegraphics[width=0.15\linewidth]{images/saliency_maps/both_correct_img_model/ILSVRC2012_val_00000994.jpeg}&
\includegraphics[width=0.15\linewidth]{images/saliency_maps/both_correct_img_model/ILSVRC2012_val_00001031.jpeg}&
\includegraphics[width=0.15\linewidth]{images/saliency_maps/both_correct_img_model/ILSVRC2012_val_00001114.jpeg}&
\includegraphics[width=0.15\linewidth]{images/saliency_maps/both_correct_img_model/ILSVRC2012_val_00000307.jpeg}\\
\\
&(a)&(b)&(c)&(d)&(e)\\
\end{tabular}
  \caption{Saliency maps when our technique gives the correct prediction and baseline approach gives incorrect label. The top row gives the saliency maps for our model and the bottom one shows the corresponding saliency maps for the model trained on imagenet alone. We can see from saliency maps that Anisotropic model has bigger saliency maps which might be the reason for the correct prediction.}
    \label{fig:correct_correct}

\end{figure*}

\section{Saliency Maps}
In Fig. \ref{fig:correct_incorrect} we show the saliency maps produced by networks trained using the combined dataset and the original ImageNet dataset. 
We use GradCam\cite{Selvaraju2016GradCAMVE} to calculate the saliency maps. We can see that Anisotropic ImageNet has saliency maps that spread out over a bigger area and that include the outlines of the objects.  This suggests that it attends less to texture and more to overall holistic shape.   In contrast, ImageNet trained models have narrower saliency maps that miss the overall shape and focus on localized regions, suggesting an attention to texture. In Fig. \ref{fig:correct_incorrect}(f-j) we show these for the case where the Anisotropic model gives the correct prediction and the ImageNet model fails.  For example in Fig. \ref{fig:correct_incorrect}(j), we see that the network trained on ImageNet alone is not focusing on the whole bird and is only focusing on the body to make the decision whereas the one trained with Anisotropic ImageNet is focusing on complete bird to make a decision.  We see a similar trend in the cases where both the models give the correct prediction (Fig. \ref{fig:correct_incorrect}(a-e)).  
In the case where Anisotropic model makes incorrect predictions and ImageNet model (Fig. \ref{fig:correct_incorrect}(k-o)) is correct we see the saliency maps are still diffused, but we fail to capture the whole object leading to incorrect predictions. 


\noindent

\bibliography{egbib}